\documentclass[runningheads]{llncs}

\usepackage{eccv}

\usepackage{eccvabbrv}
\usepackage{wrapfig}
\usepackage{graphicx}
\usepackage{booktabs}
\usepackage{subcaption}
\usepackage[accsupp]{axessibility}  % Improves PDF readability for those with disabilities.

\usepackage[pagebackref,breaklinks,colorlinks,citecolor=eccvblue]{hyperref}
\usepackage{orcidlink}

\usepackage{colortbl}
\definecolor{lightblue}{RGB}{230,240,255}

\begin{document}

\title{RefAlign: Representation Alignment for Reference-to-Video Generation} 


% Second Author\inst{2,3}\orcidlink{1111-2222-3333-4444} \and
% Third Author\inst{3}\orcidlink{2222--3333-4444-5555}}

\authorrunning{L.~Wang et al.}

\institute{\textsuperscript{1}{PCA Lab, VCIP, College of Computer Science, Nankai University, China} \\\textsuperscript{2} {Baidu Inc., China} \\ \textsuperscript{3} {PCA Lab, School of Intelligence Science and Technology, Nanjing University, China} \\ \textsuperscript{4} {College of Artificial Intelligence, Jilin University, China}
\\
% }\\@
 % \\ @outlook.com
\texttt{\small \{scitop1998\}@gmail.com},\,
\texttt{\small \{songyuxinbb\}@outlook.com},\,
\texttt{\small \{yaxing,csjyang\}@nankai.edu.cn}\\
Code: \url{https://github.com/gudaochangsheng/RefAlign}}

\author{Lei Wang\textsuperscript{1,2*,$^\ddagger$}\orcidlink{0000-0001-9356-2042},\, Yuxin Song\textsuperscript{2, $^\ddagger$}\orcidlink{0009-0002-4154-7056},\,  Ge Wu\textsuperscript{1}\orcidlink{0009-0008-3011-091X},\, Haocheng Feng\textsuperscript{2}\orcidlink{0000-0002-7567-3053},\, Hang Zhou\textsuperscript{2}\orcidlink{0000-0002-2616-923X},\ Jingdong Wang\textsuperscript{2}\orcidlink{0000-0002-4888-4445},\ Yaxing Wang\textsuperscript{4}$^\dagger$,\, Jian Yang\textsuperscript{1,3}$^\dagger$\orcidlink{0000-0003-4800-832X}
% \vspace{-8mm}
}

\maketitle

\renewcommand{\thefootnote}{$\dagger$} 
\footnotetext{Corresponding authors. *Interns in Baidu Inc. $^\ddagger$ Equal contribution}

\begin{center}
    \centering
    % \vspace{-6mm}
    \captionsetup{type=figure}
    \includegraphics[width=\linewidth]{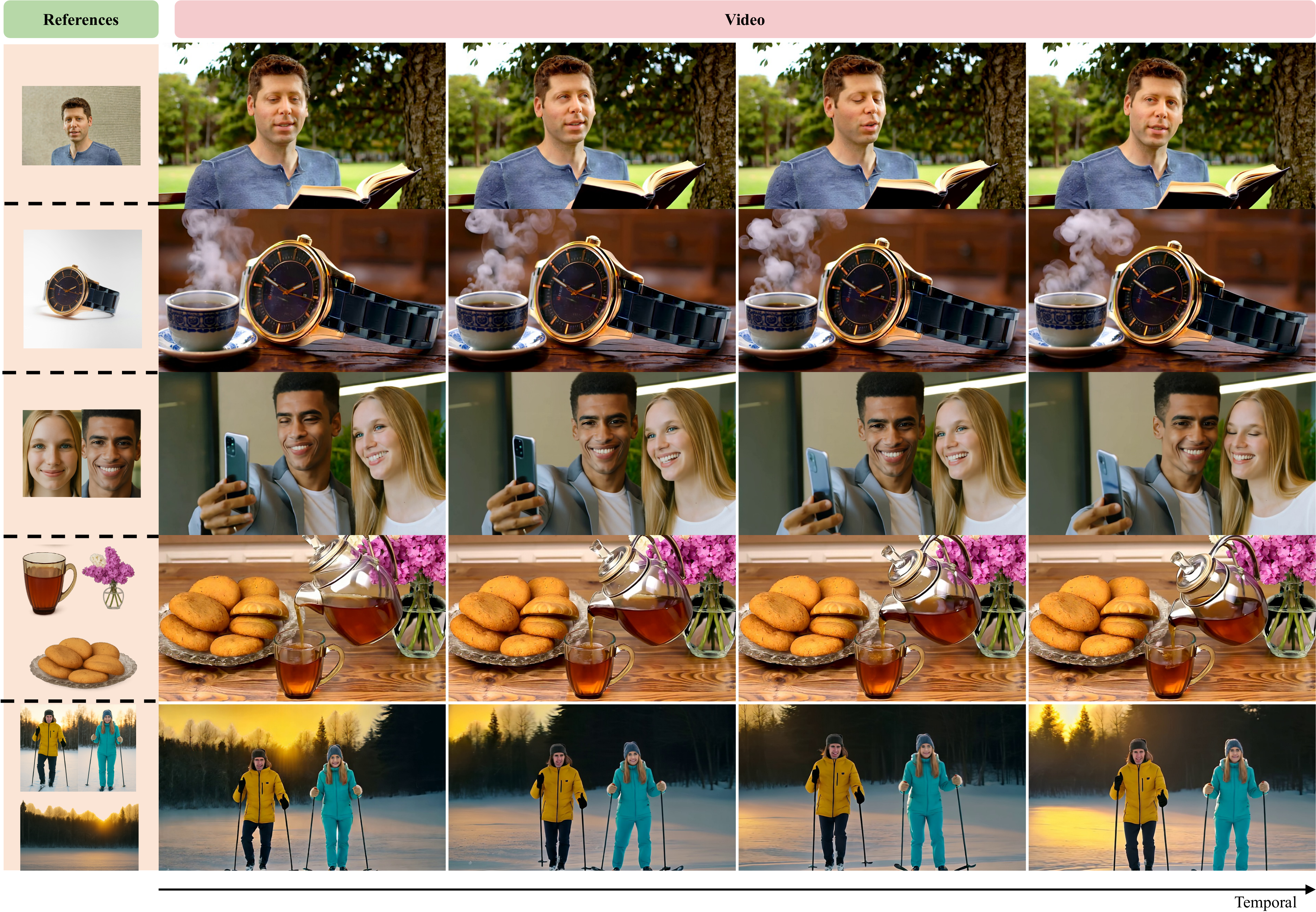}
    % \vspace{-3mm}
    \captionof{figure}{Reference-to-video generation using our proposed method, RefAlign.}\label{fig:abstract-fig}
    % \vspace{-5mm}
\end{center}%

\begin{abstract}
  Reference-to-video (R2V) generation is a controllable video synthesis paradigm that constrains the generation process using both text prompts and reference images, enabling applications such as personalized advertising and virtual try-on. In practice, existing R2V methods typically introduce additional high-level semantic or cross-modal features alongside the VAE latent representation of the reference image and jointly feed them into the diffusion Transformer (DiT). These auxiliary representations provide semantic guidance and act as implicit alignment signals, which can partially alleviate pixel-level information leakage in the VAE latent space. However, they may still struggle to address copy--paste artifacts and multi-subject confusion caused by modality mismatch across heterogeneous encoder features. In this paper, we propose RefAlign, a representation alignment framework that explicitly aligns DiT reference-branch features to the semantic space of a visual foundation model (VFM). The core of RefAlign is a reference alignment loss that pulls the reference features and VFM features of the same subject closer to improve identity consistency, while pushing apart the corresponding features of different subjects to enhance semantic discriminability. This simple yet effective strategy is applied only during training, incurring no inference-time overhead, and achieves a better balance between text controllability and reference fidelity. Extensive experiments on the OpenS2V-Eval benchmark demonstrate that RefAlign outperforms current state-of-the-art methods in TotalScore, validating the effectiveness of explicit reference alignment for R2V tasks.

\end{abstract}

\section{Introduction}
\label{sec:intro}

Diffusion models have driven rapid advances in video generation. Representative commercial systems (e.g., Sora)~\cite{brooks2024video,bao2024vidu,team2025kling} and open-source models~\cite{wan2025wan,yangcogvideox,wu2025hunyuanvideo} now achieve high-fidelity text-to-video (T2V) and image-to-video (I2V) synthesis. However, T2V relies solely on the prompt and offers limited fine-grained control (e.g., identity and appearance), while I2V is more controllable but restricts diversity due to the reference image.

To bridge this gap, reference-to-video (R2V)~\cite{chen2025multi,huang2025conceptmaster,Liu_2025_ICCV} has attracted increasing attention. Conditioned on the prompt and the reference image, it aims to generate instruction-following videos while preserving subject identity and appearance, with applications in personalized advertising~\cite{chen2025goku,liang2025movie} and virtual try-on~\cite{nguyen2025swifttry,li2025pursuing}. A key challenge, however, is effective multi-modal conditioning during video generation.

To alleviate the aforementioned challenges, prior work~\cite{Liu_2025_ICCV,fei2025skyreels,chen2025multi} typically adopts a ``two-stream reference'' paradigm: it leverages a 3D VAE to extract reference features that provide low-level details; meanwhile, it also introduces an additional encoder to encode the reference images and supply higher-level semantic cues, which can partially mitigate the detail leakage caused by the former. However, because both reference images and textual prompts are often processed by separate, independent encoders, these methods may struggle to model complex cross-modal interactions—particularly for prompts that involve spatial relations and temporal reasoning.

\begin{figure}[t]
    \centering
    \begin{subfigure}[t]{0.495\linewidth}
        \centering
        \includegraphics[width=\linewidth,height=3.0cm,keepaspectratio]{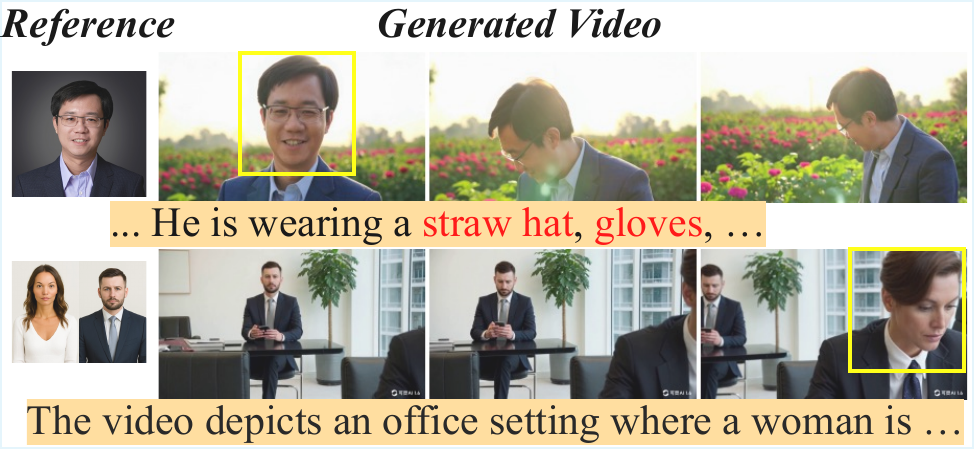}
        
        \caption{}
        % \vspace{-0.05cm}
        \label{fig:problem}
    \end{subfigure}
    \hfill
    \begin{subfigure}[t]{0.495\linewidth}
        \centering
        \includegraphics[width=\linewidth,height=3.0cm,keepaspectratio]{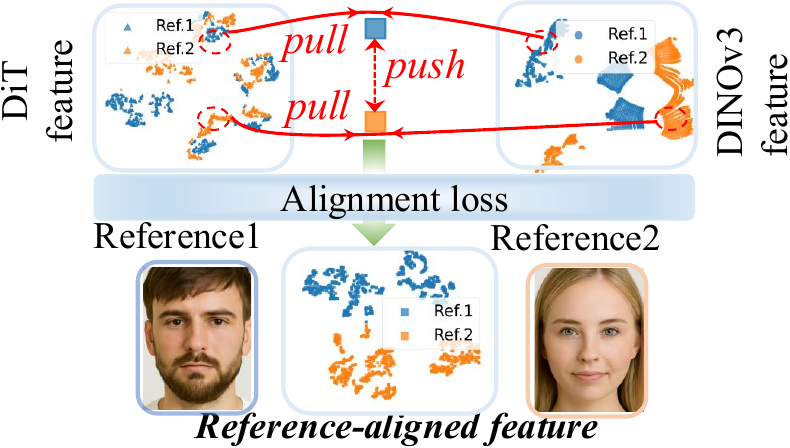}
        
        \caption{}
        % \vspace{-0.05cm}
        \label{fig:motivation}
    \end{subfigure}
    
    \begin{subfigure}[t]{\linewidth}
        \centering
        \includegraphics[width=0.95\linewidth]{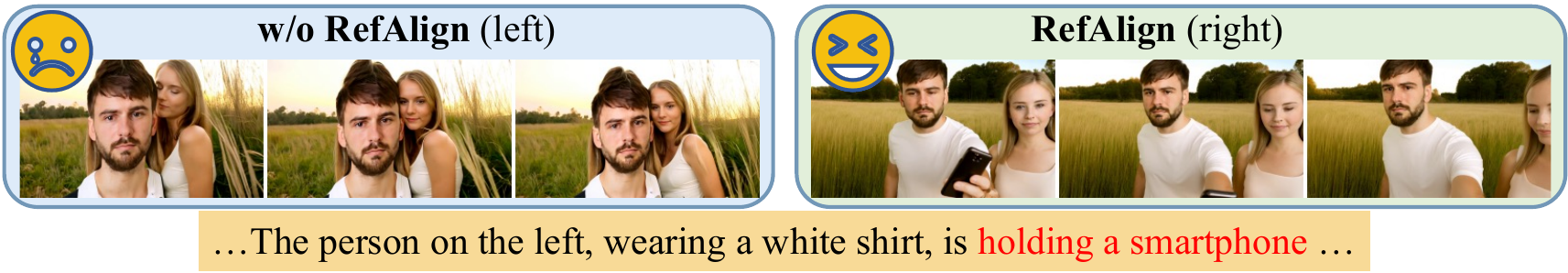} 
        % \vspace{-0.1cm}
        \caption{}
        \label{fig:case_compare} 
    \end{subfigure}
    % \vspace{-0.4cm}
    \caption{Motivation of the proposed RefAlign method. (a) The R2V task suffers from copy--paste artifacts (top) and multi-subject confusion (bottom), both generated by Kling~\cite{Kling}. (b) t-SNE~\cite{JMLR:v9:vandermaaten08a} visualization of reference feature distributions. DiT features (conditioned on VAE-encoded inputs) are highly entangled and overlap substantially across references, whereas DINOv3 features are more separable. RefAlign aligns DiT features to the DINOv3 feature space via an alignment loss, improving reference separability by \emph{pulling} same-reference features closer and \emph{pushing} different-reference features farther apart. (c) Visual comparison with and without RefAlign.
    }
    \label{fig:teaser}
    % \vspace{-0.5cm}
\end{figure}

Furthermore, some methods~\cite{deng2025cinema,li2025bindweave,pan2025id} leverage cross-modal representations from multimodal large language models (MLLM) to enhance deeper semantic interactions among multimodal inputs. However, this often incurs substantial inference overhead (e.g., introducing models such as Qwen2/2.5-VL-7B~\cite{wang2024qwen2,bai2025qwen2}). More importantly, these approaches still fundamentally rely on feeding additional semantic features together with cross-modal features into the diffusion transformer, attempting to mitigate modality mismatch via implicit alignment. \textit{Here, modality mismatch refers to the discrepancy between DiT’s internal reference representations derived from VAE latents and the external semantic reference representations injected for conditioning, making implicit alignment less effective.} Lacking explicit constraints on the alignment mechanism itself, the marginal benefit of the extra features may be limited. As a result, existing models can still suffer from copy--paste artifacts (\emph{i.e.}, overly replicating the reference image in generated videos, Fig.~\ref{fig:teaser}(\subref{fig:problem})(top)) and multi-subject confusion (\emph{i.e.}, mutual interference and blending of different subjects’ appearances, Fig.~\ref{fig:teaser}(\subref{fig:problem})(bottom)).

In this paper, we propose a simple yet effective strategy called RefAlign to mitigate the misalignment among multimodal features.
 Fig.~\ref{fig:teaser}(\subref{fig:motivation}) visualizes the latent distributions of reference images produced by different encoders. For the VAE, the features input to DiT are scattered and poorly structured, with strong cross-reference entanglement. As a result, discriminative boundaries are hard to establish. In contrast, features extracted by DINOv3~\cite{simeoni2025dinov3} demonstrate stronger separability: distributions from different reference images are more distinct, while features belonging to the same reference remain compact and consistent. Motivated by this observation, we design a vision foundation model (VFM)–guided optimization strategy for the VAE-based reference representation in DiT. We find that VFM guidance can compensate for the limited semantic expressiveness of DiT latent features, thereby alleviating copy–-paste artifacts and multi-subject confusion. Fig.~\ref{fig:teaser}(\subref{fig:case_compare}) provides qualitative evidence supporting this claim.

RefAlign’s main contribution is the proposed reference alignment (RA) loss, which aligns the DiT features of reference images into the feature space of a pretrained VFM during training. Although directly incorporating VFM features as a complement to VAE features yields performance gains—likely because additional semantic information facilitates implicit multimodal alignment—this strategy introduces an extra encoder and increases inference cost. In contrast, we find that a carefully designed RA loss plays a key role in achieving effective alignment without additional inference overhead. RA loss is designed to regularize the feature distribution while preserving the expressive capacity of DiT reference features. Moreover, to prevent overly strong similarity constraints from causing representation collapse and exacerbating multi-subject confusion, our RA loss further incorporates a negative alignment mechanism.
Finally, we systematically investigate the impact of different VFMs on alignment effectiveness.

To evaluate generation performance on the R2V task, we apply RA loss to one of the most representative video generation backbones, \textit{Wan2.1}~\cite{wan2025wan}. We conduct a comprehensive evaluation of RefAlign on the OpenS2V-Eval~\cite{yuan2025opens2v} benchmark, where it achieves state-of-the-art performance in TotalScore. Qualitative results further demonstrate clear improvements in subject fidelity and consistency. The main contributions of this paper are summarized as follows.
\begin{itemize}
    \item We propose RefAlign, the first R2V framework that regularizes reference image features by leveraging a vision foundation model. 
    \item We introduce the RA loss, which effectively alleviates both copy--paste artifacts and multi-subject confusion in R2V without introducing additional inference overhead.
    \item On the OpenS2V-Eval benchmark, RefAlign outperforms prior R2V methods, and extensive ablation studies further validate the effectiveness and necessity of the proposed design. \emph{We will make the model and code publicly available to support reproducibility and further research.}
\end{itemize}

\section{Related Work}

\subsection{Reference-to-Video Generation}

Reference-to-video (R2V) aims to synthesize high-quality videos conditioned on text and reference images. R2V has evolved from human-centric identity-preserving generation~\cite{xue2025stand,zhong2025concat,yuan2025identity,sang2025lynx} to more general objects and scenes~\cite{Liu_2025_ICCV,fei2025skyreels,li2025bindweave,zhang2025kaleido,huang2025conceptmaster}, enabling more flexible control. By reference-conditioning strategy, R2V methods fall into three categories: multi-encoder reference conditioning, reference disentanglement, and MLLM-based cross-modal guidance.

The first category introduces a semantic branch independent of the VAE encoder to impose global semantic constraints on the reference~\cite{Liu_2025_ICCV,fei2025skyreels,chen2025multi,deng2025cinema,li2025bindweave}, mitigating pixel-level leakage and copy--paste artifacts. For example, Phantom~\cite{Liu_2025_ICCV} injects fine-grained appearance cues via VAE features while anchoring subject semantics with CLIP~\cite{radford2021learning} features. 

The second category either decouples reference conditioning from video representations~\cite{zhang2025kaleido,zhou2025scaling} or explicitly aligns each reference subject with its corresponding text~\cite{deng2025magref,huang2025conceptmaster}, reducing multi-subject interference and identity confusion. For instance, Kaleido~\cite{zhang2025kaleido} adopts rotational positional encoding to distinguish image conditions from video tokens, whereas MAGREF~\cite{deng2025magref} and ConceptMaster~\cite{huang2025conceptmaster} bind subjects to their textual descriptions via masking and a decoupled attention module, respectively. 

The third category leverages MLLMs for deeper vision–text interaction to provide cross-modal relational and semantic guidance~\cite{deng2025cinema,li2025bindweave,hu2025polyvivid,hu2025hunyuancustom,pan2025id,chen2026vino}. For example, BindWeave~\cite{li2025bindweave} uses Qwen2.5-VL~\cite{bai2025qwen2} to facilitate multi-reference interaction, while PolyVivid~\cite{hu2025polyvivid} employs an internal LLaVA~\cite{liu2023visual} to embed visual identity into the text space for precise semantic alignment. In addition, VACE~\cite{jiang2025vace} is a unified model that supports both R2V and video editing via a context adapter; however, due to the difficulty of multi-task optimization, it may still struggle to fully preserve identity consistency in reference-guided generation. Unlike the above methods, we use external visual encoder features as semantic anchors and apply explicit feature alignment to pull same-subject features closer and push different-subject features apart, improving identity consistency and semantic discriminability while reducing multi-subject confusion.

\subsection{Alignment-based Method}

Recently, REPA~\cite{yurepresentation} accelerates training convergence by aligning DiT mid-block features to those of a Vision Foundation Model (VFM). Building on this, REPA-E~\cite{leng2025repa} enables end-to-end VAE tuning, while DDT~\cite{wang2025ddt} improves the alignment paradigm by decoupling the encoder and decoder. In a parallel line of work, REG~\cite{wu2025representation} and ReDi~\cite{kouzelis2025boosting} jointly model image latents and semantic signals within the diffusion process. Beyond aligning the denoiser, VA-VAE~\cite{yao2025reconstruction} extends alignment to the VAE tokenizer; ARRA~\cite{xie2025unleashing} transfers the alignment principle for autoregressive text-to-image generation; and VideoREPA~\cite{zhang2025videorepa} generalizes alignment to video generation to enhance physical consistency. In contrast, RefAlign aligns reference-conditioning features to VFM features, rather than aligning the generation target to VFM, which can strengthening conditional semantics and improving fine-tuning stability. This \emph{reference-centric} alignment approach may also inspire video editing and may promote unified modeling of understanding and generation tasks.

\section{Method}
We first briefly introduce text-to-video generation and representation alignment (REPA)~\cite{yurepresentation} in Sec.~\ref{sec.preliminary}, which form the foundation of our work. Inspired by REPA, we present the overall RefAlign pipeline in Sec.~\ref{sec.pipeline} and its core design, the reference alignment (RA) loss, in Sec.~\ref{sec:align_loss}. We further discuss the choice of target encoders for alignment in Sec.~\ref{sec:diff-encoders-alignment}. Finally, we summarize the key differences between RefAlign and REPA in Sec.~\ref{sec:relation-to-repa}.

\subsection{Preliminary}
\label{sec.preliminary}
%T2V
\noindent \textbf{Text-to-Video Generation.} Text-to-video (T2V) synthesis~\cite{wan2025wan} typically performs generative denoising modeling in a low-dimensional latent space, thereby significantly reducing computational cost. Its training objective follows the Rectified Flow (RF)~\cite{esser2024scaling} paradigm. RF defines a straight-line trajectory in this space by linearly interpolating between the noise and data latents. Therefore, the training loss function can be formulated as:
\begin{equation}
    \mathcal{L}_{\text{RF}} = \mathbb{E}_{z,\epsilon,c,t}\left[\left \| \varepsilon_{\mathrm{\Theta}}\left(z_t,c,t\right)-\left(\epsilon - z_0\right) \right \|^2_2 \right],
    \label{equ:loss_rf}
\end{equation}
where $\epsilon \sim \mathcal{N}(0, I)$ is a standard Gaussian noise sample, $z_0$ denotes the latent variable of the clean video data, $z_t$ is the noisy latent variable at timestep $t$, $c$ is the condition (e.g., prompt) and $\varepsilon_{\mathrm{\Theta}}\left(z_t,c,t\right)$ is the output predicted by the network parameterized by $\mathrm{\Theta}$.

\noindent \textbf{Representation Alignment.} Representation Alignment (REPA)~\cite{yurepresentation} is a regularization method designed to accelerate the convergence of diffusion transformers (DiT). 
It projects DiT hidden states through an MLP and aligns them with clean-image representations from a frozen pretrained vision encoder (e.g., DINOv2~\cite{oquab2024dinov2}). 
The REPA loss can be formulated as:
\begin{equation}
    \mathcal{L}_{\text{REPA}}=-\mathbb{E}\left[\frac{1}{N}\sum_{n=1}^{N}\cos{\left(f_n^*,\hat{h}_n^t
\right)}\right],
\end{equation}
where $\cos(\cdot,\cdot)$ denotes cosine similarity, $n$ is the patch index, $t$ denotes the timestep, $f^* \in \mathbb{R}^{N \times D}$ is the vision-encoder features with $N$ patch tokens of dimension $D$, and $\hat{h}^t=\mathrm{\Psi}(h^t)\in \mathbb{R}^{N \times D}$ is obtained by projecting the DiT hidden states $h^t$ through an MLP projector $\mathrm{\Psi}(\cdot)$.

%REPA
\subsection{Pipeline}
\label{sec.pipeline}

\begin{figure}[t]
    \centering
    \begin{subfigure}[t]{0.475\linewidth}
        \centering
        \includegraphics[width=\linewidth]{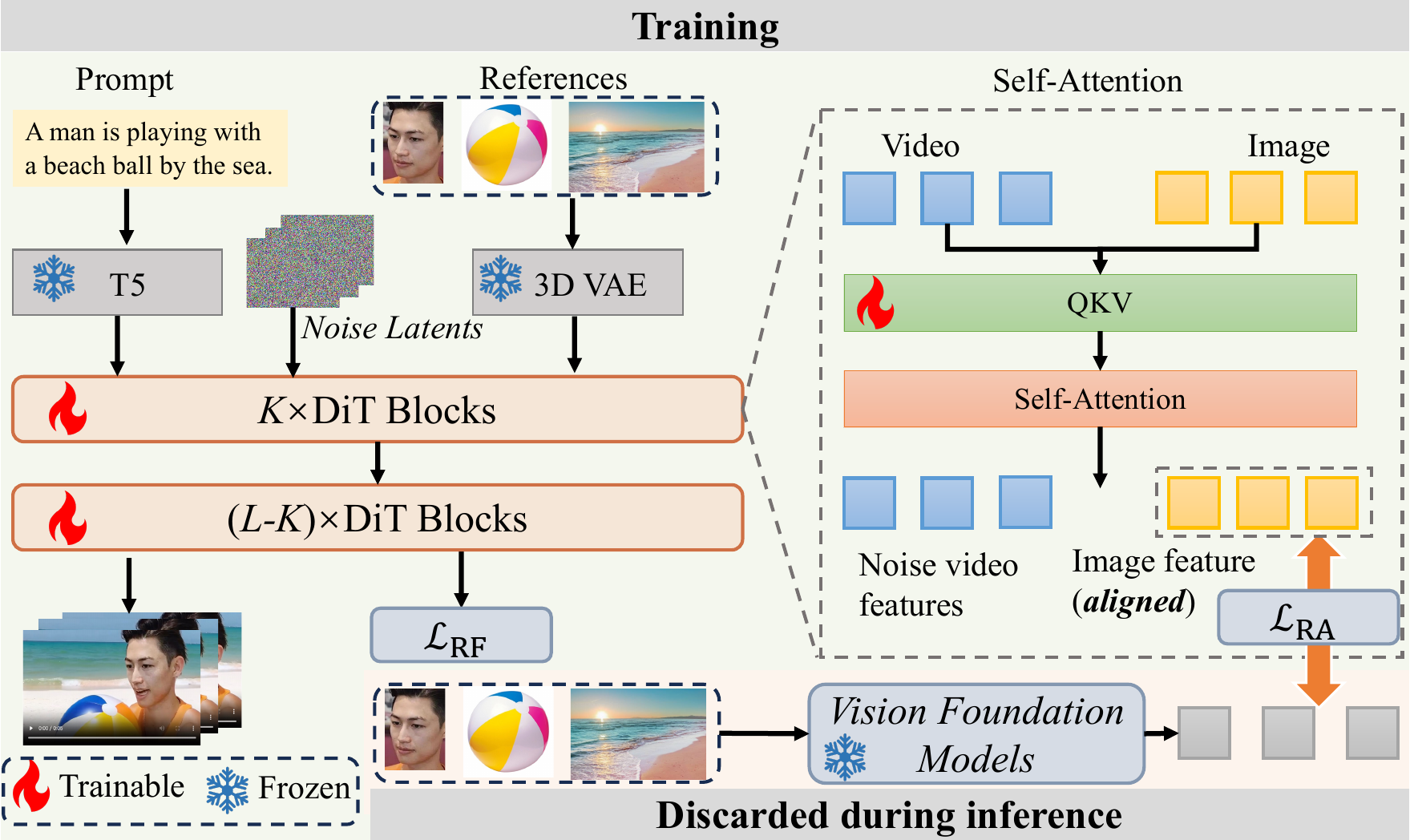}
        \caption{}
        \label{fig:refalign_pipeline}
    \end{subfigure}
    % \hfill
    \begin{subfigure}[t]{0.515\linewidth}
        \centering
        \includegraphics[width=\linewidth]{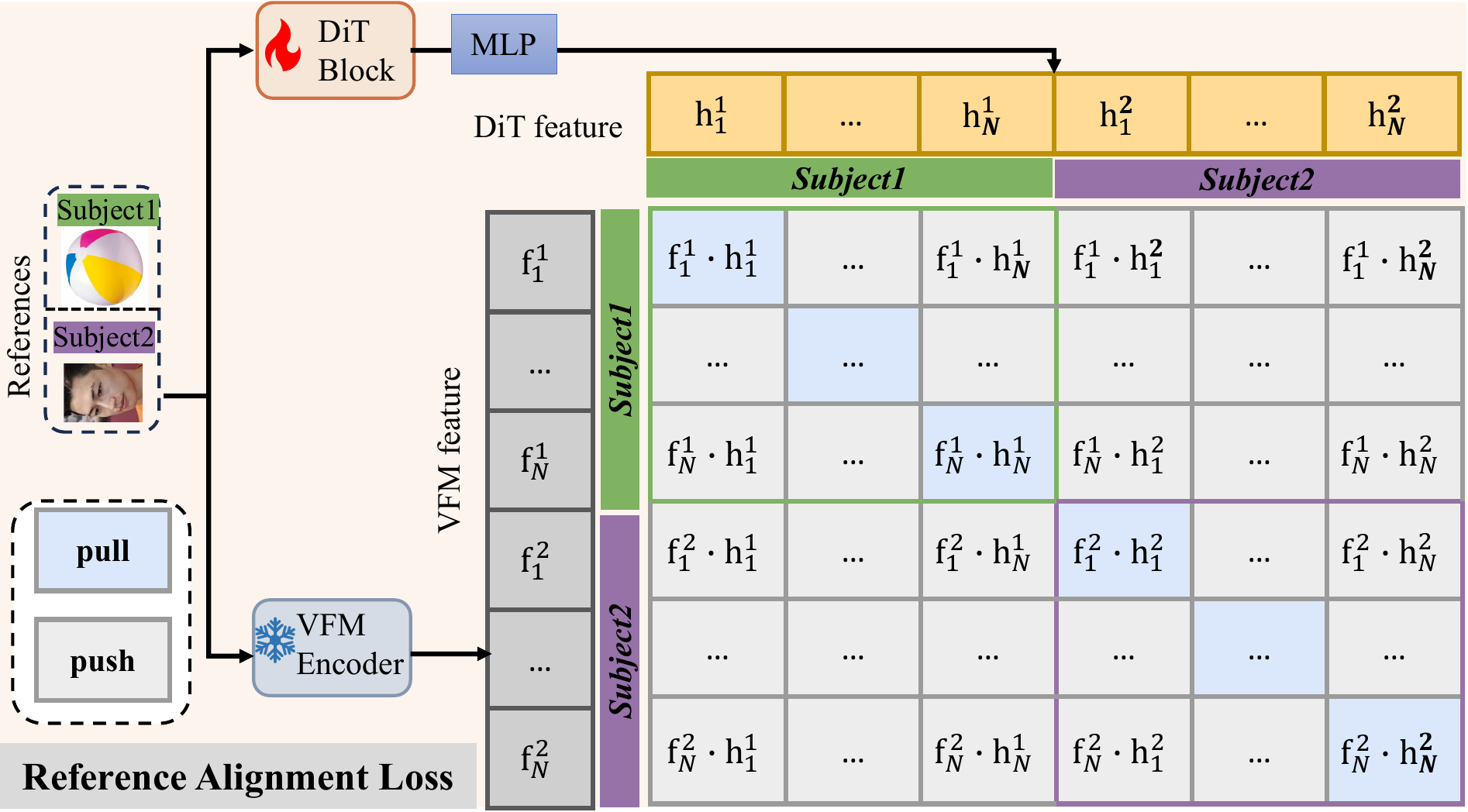}
        \caption{}
        \label{fig:refalign_loss}
    \end{subfigure}
    % \vspace{-0.3cm}
    \caption{(a) Overview of RefAlign. During training, we apply the proposed reference alignment loss $\mathcal{L}_{\mathrm{RA}}$ to intermediate features in selected DiT blocks and align them to target features extracted by a frozen vision foundation model (VFM). During inference, we discard the alignment process and the VFM. (b) Illustration of the reference alignment (RA) loss. RA loss aligns DiT reference features to their corresponding VFM teacher features by pulling matched (same-subject) pairs together and pushing mismatched (cross-subject) pairs apart, improving reference-consistent generation.}
    \label{fig:method}
    % \vspace{-0.5cm}
\end{figure}

The overall framework of RefAlign is illustrated in Fig.~\ref{fig:method}(\subref{fig:refalign_pipeline}). Our model comprises a frozen T5 encoder~\cite{raffel2020exploring} $\varepsilon_{\mathrm{T5}}$, a frozen Wan-VAE~\cite{wan2025wan} $\varepsilon_{\mathrm{VAE}}$, and a trainable diffusion transformer (DiT)~\cite{wan2025wan} parameterized by $\mathrm{\Theta}$. Given a text prompt $c_{\text{text}}$, reference images $I=\left\{I_m\right\}_{m=1}^M$, a target video $x$, and the noisy latent $z_t$ at timestep $t$, we encode the prompt, references, and target video as:
\begin{equation}
    \hat{c}_{\text{text}}=\varepsilon_{\mathrm{T5}}\left(c_{\text{text}}\right), \quad \hat{I}=\{\varepsilon_{\mathrm{VAE}}(I_m)\}_{m=1}^M, \quad z_0= \varepsilon_{\mathrm{VAE}}\left(x\right).
\end{equation}

We employ a DiT consisting of $L$ transformer blocks. Following REPA~\cite{yurepresentation}, we apply the alignment only to the intermediate \emph{reference image-token} features extracted from the first $K$ blocks (Sec.~\ref{sec:align_loss}). Importantly, all $L$ blocks are jointly optimized under the RF objective. The DiT takes $(z_t,\hat{c}_{\text{text}},\hat{I},t)$ as input and outputs 
$\varepsilon_{\mathrm{\Theta}}(z_t,\hat{c}_{\text{text}},\hat{I},t)$.
The RF objective in Eq.~(\ref{equ:loss_rf}) can be written as:
\begin{equation}
    \mathcal{L}_{\text{RF}} = \mathbb{E}_{z_0,\epsilon,t}\left[\left \| \varepsilon_{\mathrm{\Theta}}\left(z_t,\hat{c}_{\text{text}},\hat{I},t\right)-\left(\epsilon - z_0\right) \right \|^2_2 \right].
    \label{equ:rf_refalign}
\end{equation}
\noindent \textbf{Inference.} At inference, we remove the training-time VFM and MLP (Fig.~\ref{fig:method}(\subref{fig:refalign_pipeline})) to avoid extra overhead. Following Phantom~\cite{Liu_2025_ICCV} and Video Alchemist~\cite{chen2025multi}, we use an RF sampler with classifier-free guidance (CFG)~\cite{ho2022classifier}. At timestep $t$, the guided prediction is:
\begin{equation}
\begin{aligned}
% \small
\hat{\varepsilon}_{\mathrm{\Theta}}\left(z_t,\hat{c}_{\text{text}},\hat{I},t\right)
&=\varepsilon_{\mathrm{\Theta}}\left(z_t,\oslash,\oslash,t\right)
+\mu_1\Big(\varepsilon_{\mathrm{\Theta}}\left(z_t,\oslash,\hat{I},t\right)
-\varepsilon_{\mathrm{\Theta}}\left(z_t,\oslash,\oslash,t\right)\Big) \\
&\quad +\mu_2\Big(\varepsilon_{\mathrm{\Theta}}\left(z_t,\hat{c}_{\text{text}},\hat{I},t\right)
-\varepsilon_{\mathrm{\Theta}}\left(z_t,\oslash,\hat{I},t\right)\Big).
\end{aligned}
\label{equ:cfg}
\end{equation}
where $\oslash$ denotes dropping the corresponding condition (i.e., a null text embedding or a null reference embedding). $\mu_1$ and $\mu_2$ balance the reference-image and text conditions.

\subsection{Reference Alignment Loss}
\label{sec:align_loss}

By comparing the feature distributions of reference images encoded by DiT and DINOv3, we observe that DiT features are strongly coupled across references, whereas DINOv3 features are more separable. Motivated by this observation, we introduce a reference alignment (RA) loss that aligns DiT reference representations to the vision foundation model (VFM) feature space. As shown in Fig.~\ref{fig:method}(\subref{fig:refalign_pipeline}), we align the \emph{reference image-token} features produced inside the self-attention of the first $K$ DiT blocks to features extracted by a frozen VFM $\varepsilon_{\mathrm{VFM}}\left(\cdot\right)$.
Specifically, for each training sample we obtain the VFM features from $I$,
\begin{equation}
    f = \{\varepsilon_{\mathrm{VFM}}(I_m)\}_{m=1}^M, \in \mathbb{R}^{M \times N \times D}.
\end{equation}

Let $h^{(l)}=\varepsilon_{\mathrm{\Theta}^{(l)}}({I})$ be the \emph{reference image-token} features from the self-attention module of the $l$-th DiT block ($l\le K$), and we project them with an MLP $\mathrm{\Psi}_{\text{proj}}(\cdot)$ to match the VFM feature dimension:
\begin{equation}
    \hat{h}^{(l)} = \mathrm{\Psi}_{\text{proj}}\left(h^{(l)}\right), \in\mathbb{R}^{M \times N\times D}.
\end{equation}

As illustrated in Fig.~\ref{fig:method}(\subref{fig:refalign_loss}), we define the RA loss with a positive (diagonal) term and a negative (off-diagonal) term. We pull DiT reference tokens closer to the corresponding VFM tokens for the same subject:
\begin{equation}
    \mathcal{L}_{\mathrm{pos}}^{(l)}
    = \frac{1}{M}\sum_{m=1}^M\frac{1}{N}\sum_{n=1}^{N}
    \left(1-\cos\left(\hat{h}^{(l),m}_{n}, f_{n}^m\right)\right).
\end{equation}
To enforce subject-level separation, we push features from subject $m$ away from features of other subjects $m'\neq m$ using a margin $\delta$:
\begin{equation}
\mathcal{L}^{(l)}_{\mathrm{neg}}
=
\frac{1}{M(M-1)}
\sum_{m=1}^{M}
\sum_{\substack{m'=1 \\ m' \neq m}}^{M}
\frac{1}{N^{2}}
\sum_{n=1}^{N}
\sum_{n'=1}^{N}
\left[\,\delta - \left(1-\cos\left(\hat{h}^{(l),m}_{n}, f_{n^{\prime}}^{m^\prime}\right)\right)\right]_{+},
\end{equation}
where $[x]_+=\max(x,0)$. Then, we average over the first $K$ blocks and combine the two terms:
\begin{equation}
\mathcal{L}_{\mathrm{RA}}
=
\frac{1}{K}\sum_{l=1}^{K}
\left(
\mathcal{L}^{(l)}_{\mathrm{pos}}
+
\lambda\,\mathcal{L}^{(l)}_{\mathrm{neg}}
\right),
\label{euq:ra}
\end{equation}
where $\lambda$ controls the weight of the negative term. \textbf{Special case.} When $M=1$ (i.e., only one reference), the inter-subject negative term is undefined and we set $\mathcal{L}^{(l)}_{\mathrm{neg}}=0$, so $\mathcal{L}_{\mathrm{RA}}$ reduces to the positive alignment loss. Finally, combining Eq.~\eqref{equ:rf_refalign} and Eq.~\eqref{euq:ra}, we optimize our model with the following overall objective:
\begin{equation}
\mathcal{L}=\mathcal{L}_{\text{RF}}+\eta\,\mathcal{L}_{\text{RA}},
\label{equ:total}
\end{equation}
where $\eta$ is a scalar hyperparameter balancing RA and RF losses.

\subsection{Alignment Using Different Encoders}
\label{sec:diff-encoders-alignment}
In RefAlign, $\mathcal{E}_{\text{VFM}}$ aligns the intermediate features of DiT’s reference branch to establish a stable subject anchor. Therefore, the representational characteristics of $\mathcal{E}_{\text{VFM}}$ strongly influence the behavior of the alignment constraint. Below, we discuss the differences induced by three types of encoders.

\noindent \textbf{DINOv3.} DINOv3 provides self-supervised patch representations with strong instance-level discriminability~\cite{simeoni2025dinov3,yurepresentation}. When used as the alignment target, it encourages the reference branch to preserve identity-relevant cues while being less sensitive to appearance variations such as background, illumination, and pose. Its limitation is the lack of language supervision, resulting in relatively weaker high-level semantic constraints. \emph{Therefore, using $\mathcal{E}_{\text{VFM}}^{\text{DINOv3}}$ steers RefAlign to emphasize intra-subject representation consistency and inter-subject separability, helping mitigate the effects of appearance variations and cross-subject interference on the reference-branch representations.}
 
\noindent \textbf{SigLIP2.} SigLIP2 is trained using contrastive learning, which encourages patch representations to be consistent with global semantic separability~\cite{tschannen2025siglip}. As a result, local cues are often consolidated into more stable category- or attribute-level features. However, when the subject and background are highly co-occurring, this global discriminative bias can also encode background co-occurrence patterns as informative signals. \emph{Therefore, when using $\mathcal{E}_{\text{VFM}}^{\text{SigLIP2}}$ as the alignment target, RefAlign tends to emphasize consistency along semantic-attribute dimensions: attribute cues that persist in the reference images and align with the prompt semantics are more likely to be amplified in the reference-branch representations.}

\noindent \textbf{Qwen2.5-VL.} Qwen2.5-VL vision tokens are strongly coupled with textual semantics through multimodal pre-training, making them well-suited for providing cross-modal semantic supervision~\cite{bai2025qwen2}. When used as the alignment target, the reference branch is more likely to be pulled toward a prompt-consistent semantic direction (in contrast to SigLIP2, which tends to emphasize category/attribute-level semantics). A limitation is that multimodal fusion and sequence structuring can weaken the stability of local spatial correspondences, and the model is heavier (7B), incurring higher training cost and VRAM pressure. \emph{Therefore, when using $\mathcal{E}_{\text{VFM}}^{\text{Qwen2.5-VL}}$ as the alignment target, RefAlign places more emphasis on contextualized compositional semantics, encouraging the reference-branch representation to organize reference cues into concept-level features aligned with the prompt semantics.}

Overall, RefAlign tends to benefit more from supervision that is \textbf{spatially consistent, identity-sensitive, and robust to appearance variations}. DINOv3 may align better with this form of supervision; by contrast, SigLIP2 tends to emphasize semantic discrimination and Qwen2.5-VL leans toward multimodal semantic fusion, making their alignment signals more likely to deviate from pure ``subject anchoring''.

\subsection{Relationship to REPA}
\label{sec:relation-to-repa}

\begin{figure}[htbp]
  \centering
    % \vspace{-0.6cm}
  \begin{subfigure}[t]{0.36\textwidth}
    \centering
    \includegraphics[width=\linewidth]{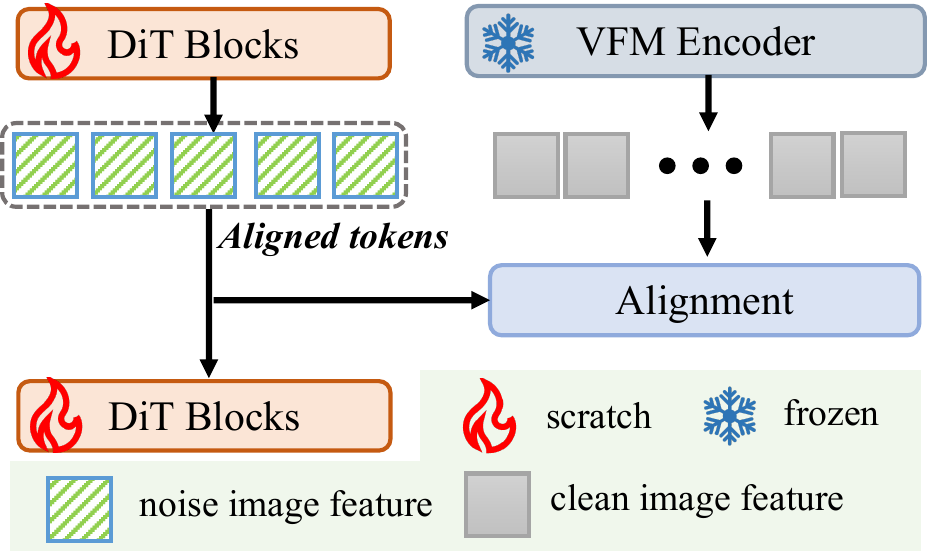}
    \caption{REPA Training}
    \label{fig:repa-train}
  \end{subfigure}
  % \hfill
  \begin{subfigure}[t]{0.37\textwidth}
    \centering
    \includegraphics[width=\linewidth]{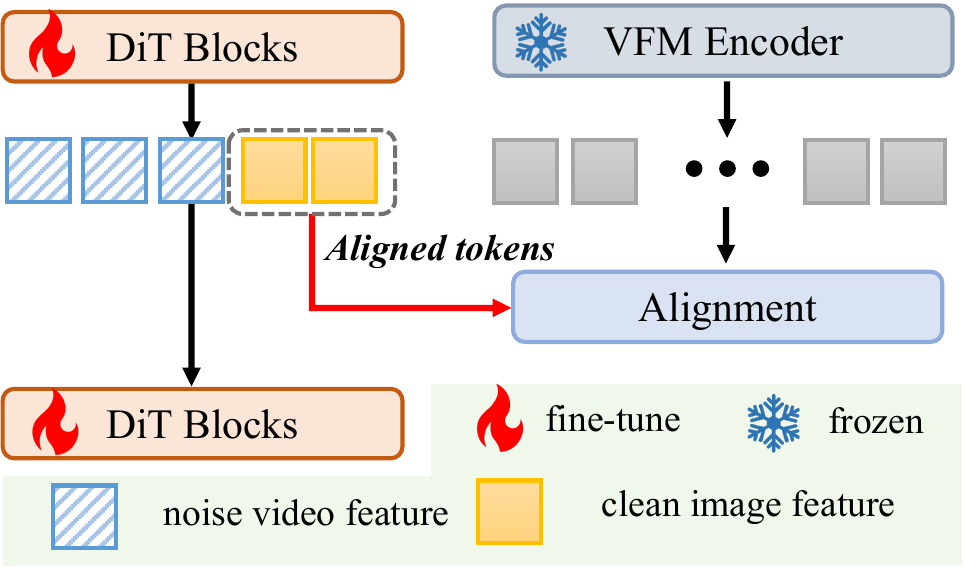}
    \caption{RefAlign Training}
    \label{fig:refalign-train}
  \end{subfigure}
% \vspace{-0.3cm}
  \caption{Training comparison between REPA and RefAlign. (a) REPA: Trained from scratch, aligning noisy generation targets with clean VFM features to accelerate DiT convergence. (b) RefAlign: Fine-tuned from \textit{Wan2.1}~\cite{wan2025wan} initialization, aligning clean reference-branch image features with clean VFM features to optimize reference representations and improve reference controllability.}
  \label{fig:repa-vs-ref}
  % \vspace{-0.5cm}
\end{figure}

Our RefAlign and REPA both leverage representations from VFM to assist DiT training, but they differ fundamentally in motivation, objective, and mechanism (as shown in Fig.~\ref{fig:repa-vs-ref}). \textbf{1)} \textbf{Different motivations}: REPA uses VFM to regularize DiT’s \emph{target} representations, easing semantic learning from noise. However, RefAlign regularizes the \emph{reference condition} with VFM to mitigate VAE-induced pixel-level leakage and copy--paste artifacts, improving reference controllability. \textbf{2)} \textbf{Different objectives}: REPA targets training \emph{from scratch}, using semantic regularization to speed convergence. By contrast, RefAlign targets \emph{fine-tuning} pretrained video generators (e.g., \textit{Wan2.1}~\cite{wan2025wan}), balancing text and reference conditions while largely preserving generative quality. \textbf{3)} \textbf{Different task suitability}: In multi-reference settings, REPA’s one-to-one alignment can be overly hard, potentially collapsing different reference representations toward an average and thus reducing discriminability and increasing subject confusion (see Tab.~\ref{tab:compont-abla} and Fig.~\ref{fig:vis-abla}). RefAlign instead enforces inter-reference separability via cross-subject discrimination (e.g., $\mathcal{L}_{\text{neg}}$), making it well suited to multi-reference-to-video generation. \textbf{4)} \textbf{Different alignment representations}: REPA pulls \emph{noisy} representations toward VFM, making it more suitable for training from scratch. In contrast, RefAlign aligns the \emph{clean} DiT reference-image representations to VFM, reducing interference with the generative backbone for stable fine-tuning.

\begin{table}[t]
\centering
\caption{Quantitative comparison of RefAlign and other methods on zero-shot OpenS2V-Eval results.  $\uparrow$ denotes that higher is better.}
\resizebox{0.999\linewidth}{!}{
\begin{tabular}{lcccccccc}
\toprule
\textbf{Method} & \textbf{TotalScore}$\uparrow$ & \textbf{Aesthetics}$\uparrow$ & \textbf{MotionSmoothness}$\uparrow$ & \textbf{MotionAmplitude}$\uparrow$ & \textbf{FaceSim}$\uparrow$ & \textbf{GmeScore}$\uparrow$ & \textbf{NexusScore}$\uparrow$ & \textbf{NaturalScore}$\uparrow$ \\
\midrule
\multicolumn{9}{c}{\textbf{\textit{Closed-source}}} \\
\midrule
\midrule
Vidu2.0~\cite{bao2024vidu}        & 51.95\% & 41.48\% & 90.45\% & 13.52\% & 35.11\% & 67.57\% & 43.37\% & 65.88\% \\
Pika2.1~\cite{Pika}        & 51.88\% & 46.88\% & 87.06\% & 24.71\% & 30.38\% & 69.19\% & 45.40\% & 63.32\% \\
Kling1.6~\cite{Kling}       & 56.23\% & 44.59\% & 86.93\% & 41.60\% & 40.10\% & 66.20\% & 45.89\% & 74.59\% \\
Saber-14B~\cite{zhou2025scaling}          & 57.91\% & 42.42\% & 96.12\% & 21.12\% & 49.89\% & 67.50\% & 47.22\% & 72.55\% \\
\midrule
\multicolumn{9}{c}{\textbf{\textit{Open-source}}} \\
\midrule
\midrule
VACE-1.3B~\cite{jiang2025vace}      & 49.89\% & 48.24\% & 97.20\% & 18.83\% & 20.57\% & 71.26\% & 37.91\% & 65.46\% \\
VACE-P1.3B~\cite{jiang2025vace}     & 48.98\% & 47.34\% & 96.80\% & 12.03\% & 16.59\% & 71.38\% & 40.19\% & 64.31\% \\
Phantom-1.3B~\cite{Liu_2025_ICCV}   & 54.89\% & 46.67\% & 93.30\% & 14.29\% & 48.56\% & 69.43\% & 42.48\% & 62.50\% \\
\midrule
\rowcolor{lightblue} \textbf{RefAlign-1.3B} & \textbf{56.30}\% & 42.96\% & 94.74\% & \textbf{20.74}\% & \textbf{53.06}\% & 66.85\% & \textbf{43.97}\% & \textbf{66.25}\%  \\
\midrule
VACE-14B~\cite{jiang2025vace}       & 57.55\% & 47.21\% & 94.97\% & 15.02\% & 55.09\% & 67.27\% & 44.08\% & 67.04\% \\
SkyReels-A2-P14B~\cite{fei2025skyreels} & 52.25\% & 39.41\% & 87.93\% & 25.60\% & 45.95\% & 64.54\% & 43.75\% & 60.32\% \\
Phantom-14B~\cite{Liu_2025_ICCV}    & 56.77\% & 46.39\% & 96.31\% & 33.42\% & 51.46\% & 70.65\% & 37.43\% & 69.35\% \\
MAGREF-480P~\cite{deng2025magref}    & 52.51\% & 45.02\% & 93.17\% & 21.81\% & 30.83\% & 70.47\% & 43.04\% & 66.90\% \\
VINO~\cite{chen2026vino}           & 57.85\% & 45.92\% & 94.73\% & 12.30\% & 52.00\% & 69.69\% & 42.67\% & 71.99\% \\
Kaleido~\cite{zhang2025kaleido}        & 56.59\% & 46.15\% & 98.11\% & 12.67\% & 34.56\% & 67.51\% & 41.70\% & 82.18\% \\
BindWeave~\cite{li2025bindweave}      & 57.61\% & 45.55\% & 95.90\% & 13.91\% & 53.71\% & 67.79\% & 46.84\% & 66.85\% \\
\midrule
\rowcolor{lightblue} \textbf{RefAlign-14B} & \textbf{60.42}\% & 46.84\% & 97.61\% & 22.48\% & \textbf{55.23}\% & 68.32\% & \textbf{48.52}\% & 73.63\%  \\

\bottomrule
\end{tabular}
}
\label{tab:main}
% \vspace{-0.5cm}
\end{table}

\section{Experiment}
\label{sec:blind}

\subsection{Experiment Setting}
\noindent \textbf{Datasets and Evaluation.} RefAlign is fine-tuned on a high-quality subset (360K samples) curated from OpenS2V-5M~\cite{yuan2025opens2v} and Phantom-Data~\cite{chen2025phantom-data}. The subset consists of image–text–video triplets, with a regular-pair to cross-pair ratio of 6:4. To validate effectiveness, we generate 180 videos on the OpenS2V-Eval~\cite{yuan2025opens2v} benchmark. We report Aesthetics (visual quality), MotionSmoothness (motion continuity), MotionAmplitude (motion magnitude), FaceSim (face fidelity), NexusScore (subject consistency), NaturalScore (naturalness), and Gme\linebreak Score (video–text alignment).

\noindent \textbf{Implementation Details.} RefAlign is fine-tuned on the \textit{Wan2.1}~\cite{wan2025wan} T2V backbone. The MLP for \(\mathcal{L}_{\mathrm{RA}}\) has two linear layers with SiLU, and its parameters are not shared across DiT blocks. The model is trained for 3000 iterations with AdamW~\cite{loshchilov2017decoupled} ($\beta_1=0.9$, $\beta_2=0.999$, $\text{weight decay}=0.01$), a learning rate of 5$e$-5, and a global batch size of 128. To mitigate copy--paste artifacts, we apply data augmentation to the reference images in regular pairs (e.g., random rotation, scaling, horizontal flip, affine transformations (including shear), Gaussian blur, and color jitter). We set $\lambda$ and $\eta$ in Eqs.~(\ref{euq:ra}) and~(\ref{equ:total}) to 1.0. During inference, we use a 50-step Euler sampler, with $\mu_1$ and $\mu_2$ in Eq.~(\ref{equ:cfg}) set to 5.0 and 7.5, respectively.

\noindent \textbf{Baselines.} We choose \textit{closed-source} models (e.g., Vidu~\cite{bao2024vidu}, Pika~\cite{Pika}, Kling~\cite{Kling}, and Saber~\cite{zhou2025scaling}) and \textit{open-source} methods (e.g., VACE~\cite{jiang2025vace}, SkyReels-A2~\cite{fei2025skyreels}, Phantom~\cite{Liu_2025_ICCV}, MAGREF~\cite{deng2025magref}, VINO~\cite{chen2026vino}, Kaleido~\cite{zhang2025kaleido}, and BindWeave~\cite{li2025bindweave}) as baselines for comparison with RefAlign.

\begin{figure}[htbp]
  \centering
  % \vspace{-0.5cm}
  \includegraphics[width=\linewidth]{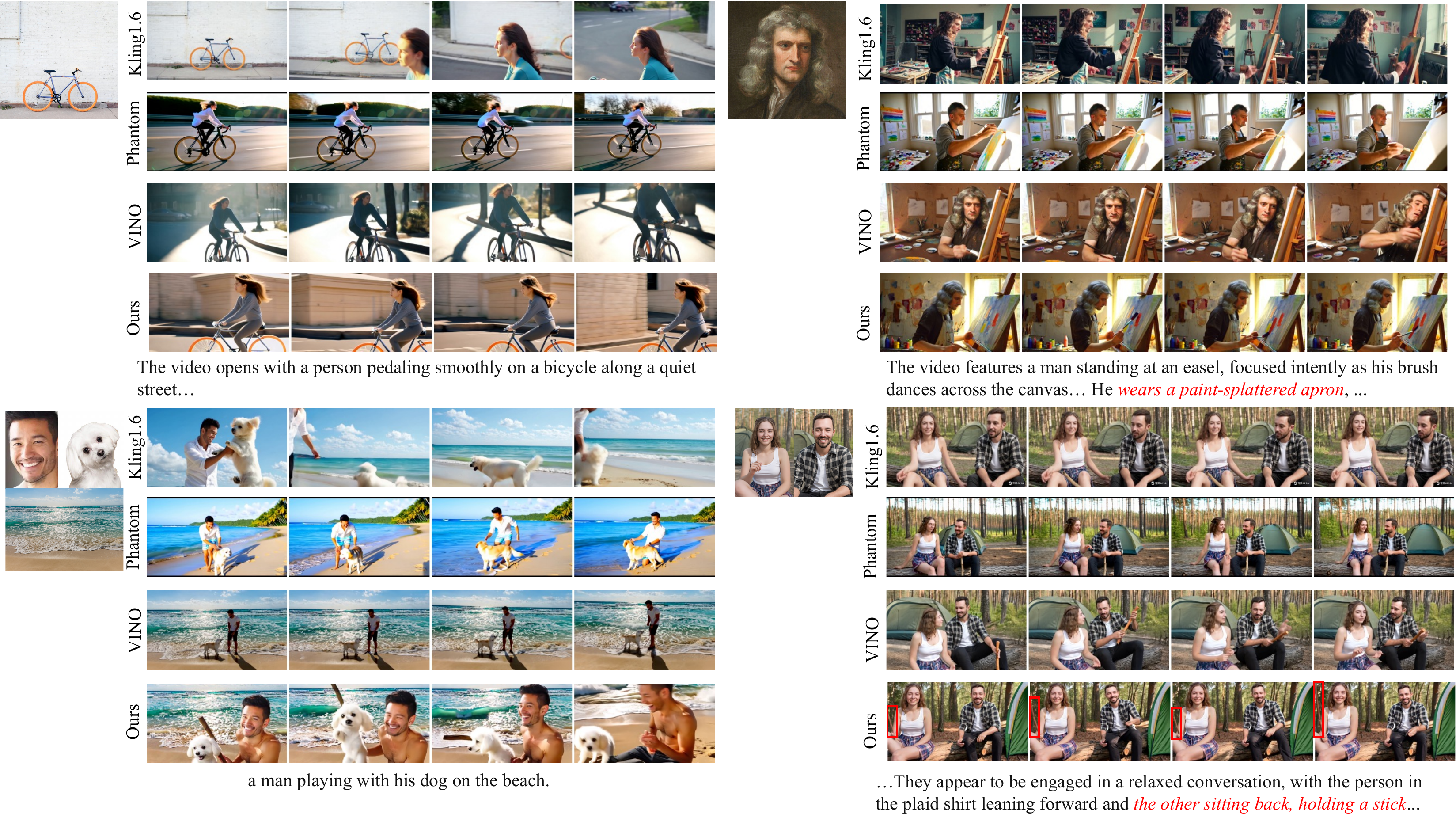}
  % \vspace{-0.6cm}
  \caption{Qualitative results. We compare RefAlign with three representative methods, namely Kling1.6~\cite{Kling}, Phantom~\cite{Liu_2025_ICCV}, and VINO~\cite{chen2026vino}.}
  \label{fig:vis-main}
  \vspace{-1.0cm}
\end{figure}

\subsection{Quantitative Results}
Tab.~\ref{tab:main} reports the R2V generation performance of RefAlign, open-source baselines, and closed-source baselines on the OpenS2V-Eval benchmark. For the 1.3B model, RefAlign achieves the state-of-the-art (SOTA) TotalScore, indicating a better overall trade-off across evaluation metrics than competing methods. Specifically, RefAlign obtains the best results on NexusScore and FaceSim, demonstrating its clear advantage in subject consistency and fidelity. Meanwhile, RefAlign also delivers competitive performance on NaturalScore. For the 14B model, RefAlign likewise achieves the SOTA TotalScore, further validating its effectiveness across different model scales. To the best of our knowledge, RefAlign is the first model to achieve a TotalScore of 60.42\% among existing publicly reported results.

\subsection{Qualitative Results}
Fig.~\ref{fig:vis-main} presents qualitative comparisons between RefAlign and existing SOTA R2V methods. Overall, RefAlign achieves a better balance between preserving the appearance consistency of the reference subject and following the text prompt, producing videos with more stable temporal coherence and smoothness. In contrast, Kling1.6 shows noticeable copy--paste artifacts in some examples (top-left), and the subject fidelity may degrade in certain scenarios (bottom-left); Phantom exhibits similar issues. In addition, both methods appear to underutilize the background reference image in some cases, leading to reduced background consistency (bottom-left). The strongest open-source baseline, VINO, still shows erroneous copying of clothing textures in some results (top-right), and its instruction following becomes weaker under some prompts (bottom-right).

\subsection{Ablation Study}

To validate the effectiveness of the RA loss, we conduct a systematic ablation study from three aspects: loss design, alignment depth, and the alignment encoder. Notably, all experiments are conducted under a unified setting of \textbf{\emph{1800 training iterations}} to ensure a fair comparison. The baseline configuration (w/o $\mathcal{L}_{\mathrm{RA}}$) only encodes the reference image with the VAE and feeds it into the DiT, without introducing the RA loss.
\begin{table}[t]
\centering

\caption{Ablation study on the impact of the RA loss design in RefAlign on the OpenS2V-Eval. \textbf{Config D} adopts a dual-encoder input, feeding both VAE and DINOv3 features into DiT simultaneously.}
\resizebox{0.999\linewidth}{!}{
\begin{tabular}{lcccccccc}
\toprule
Configuration & TotalScore$\uparrow$ & Aesthetics$\uparrow$ & MotionSmoothness$\uparrow$ & MotionAmplitude$\uparrow$ & FaceSim$\uparrow$ & GmeScore$\uparrow$ & NexusScore$\uparrow$ & NaturalScore$\uparrow$ \\
\midrule
A: $\mathcal{L}_{\mathrm{RA}}$ & \textbf{55.73\%} & 41.53\% & 90.00\% & 29.53\% & 53.15\% & 67.54\% & 46.23\% & 62.96\% \\
B: $\mathcal{L}_{\mathrm{RA}}$(w/o $\mathcal{L}_{\mathrm{neg}}$) & 51.75\% & 41.12\% & 91.80\% & 17.67\% & 48.67\% & 61.31\% & 46.61\% & 53.75\% \\
C: w/o $\mathcal{L}_{\mathrm{RA}}$ & 49.93\% & 36.46\% & 88.97\% & 28.62\% & 68.45\% & 58.83\% & 38.63\% & 40.46\% \\
D: VAE + DINOv3 & 52.15\% & 37.14\% & 87.98\% & 39.26\% & 35.11\% & 67.58\% & 45.78\% & 66.06\% \\
\bottomrule
\end{tabular}
}
\label{tab:compont-abla}
% \vspace{-0.5cm}
\end{table}

\begin{figure}[htbp]
% \vspace{-0.5cm}
  \centering
  \includegraphics[width=0.83\linewidth]{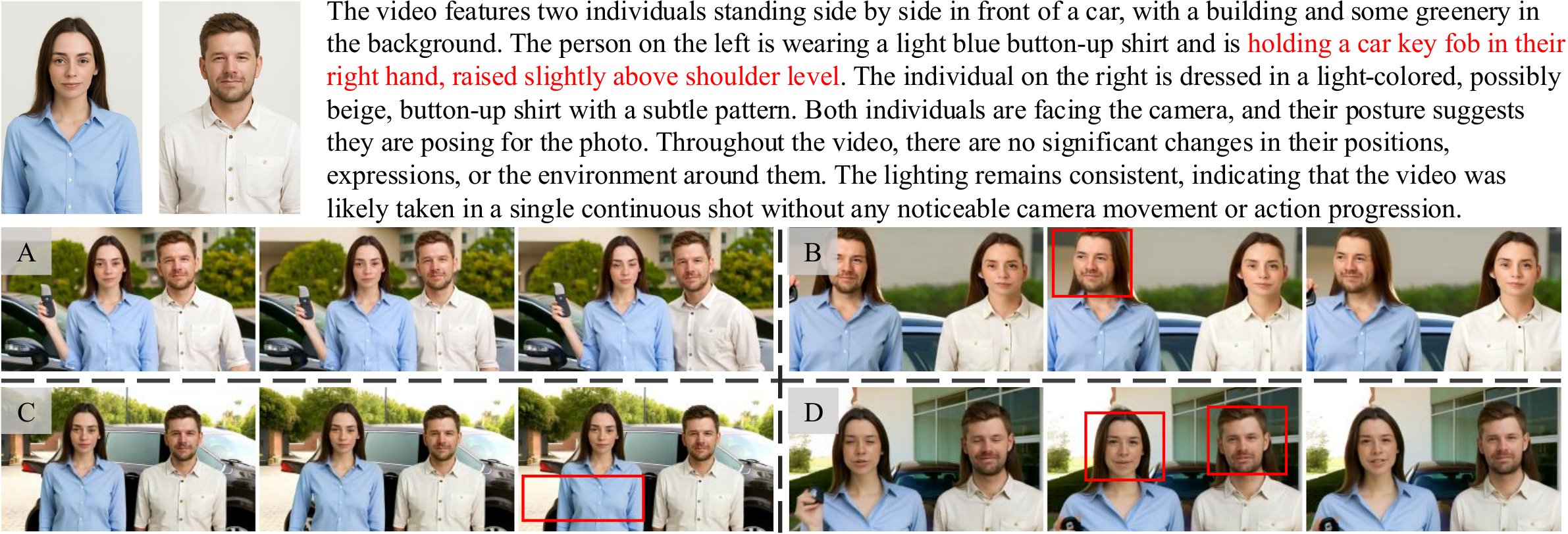}
  % \vspace{-0.1cm}
  \caption{Qualitative Ablation Study. RA loss (A) improves reference fidelity and instruction following. Removing the negative loss (B) leads to multi-reference confusion; removing RA loss (C) causes copy--paste artifacts; using DINOv3 features as input (D) reduces subject fidelity.}
  \label{fig:vis-abla}
  \vspace{-0.5cm}
\end{figure}

\noindent\textbf{Effectiveness of RA loss.} To validate the design of the proposed RA loss, we conduct an ablation study with different configurations. As shown in Tab.~\ref{tab:compont-abla}, configuration A achieves the best TotalScore (55.73\%), indicating a favorable trade-off. Configuration B lowers TotalScore mainly due to drops in FaceSim and NaturalScore, showing $\mathcal{L}_{\text{neg}}$ is important for suppressing incorrect alignments and improving naturalness. Without RA loss (configuration C), TotalScore drops sharply (55.73\%$\rightarrow$49.93\%) while FaceSim increases markedly, indicating RA loss mitigates copy--paste artifacts. Configuration D improves over C (52.15\% vs. 49.93\%), but still trails A, suggesting direct alignment to DINOv3 is more effective than using DINOv3 as an extra input, likely due to reduced modality mismatch and more stable fine-tuning.

The qualitative comparisons in Fig.~\ref{fig:vis-abla} visually support the above analysis. Configuration B shows clear confusion in gender-related attributes; configuration C retains high facial fidelity but fails to follow ``holding a car key''; configuration D better follows the action instruction, yet facial fidelity drops and slight identity confusion emerges. Overall, configuration A offers a more stable trade-off between instruction following and identity consistency.

\begin{figure}[t]
    \centering
    \begin{subfigure}[t]{0.32\columnwidth}
        \centering
        \includegraphics[width=\linewidth]{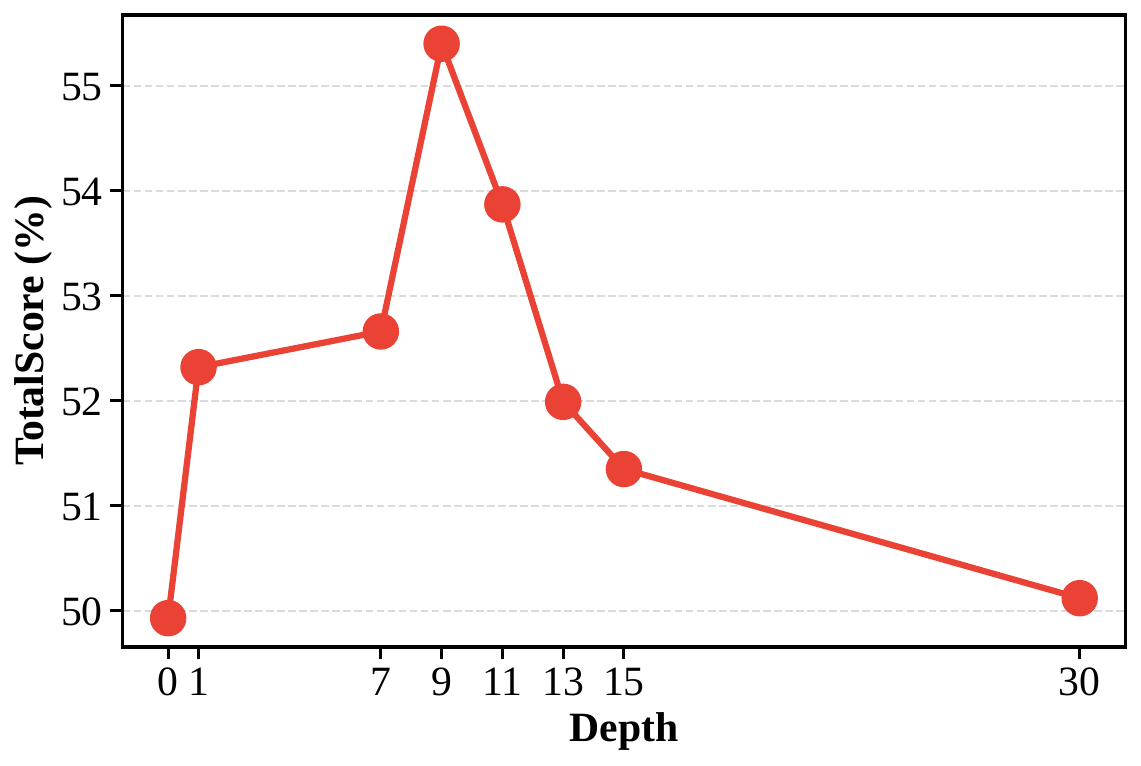}
        \vspace{-0.5cm}
        \caption{}
        \label{fig:depth_totalscore}
    \end{subfigure}\hfill
    \begin{subfigure}[t]{0.32\columnwidth}
        \centering
        \includegraphics[width=\linewidth]{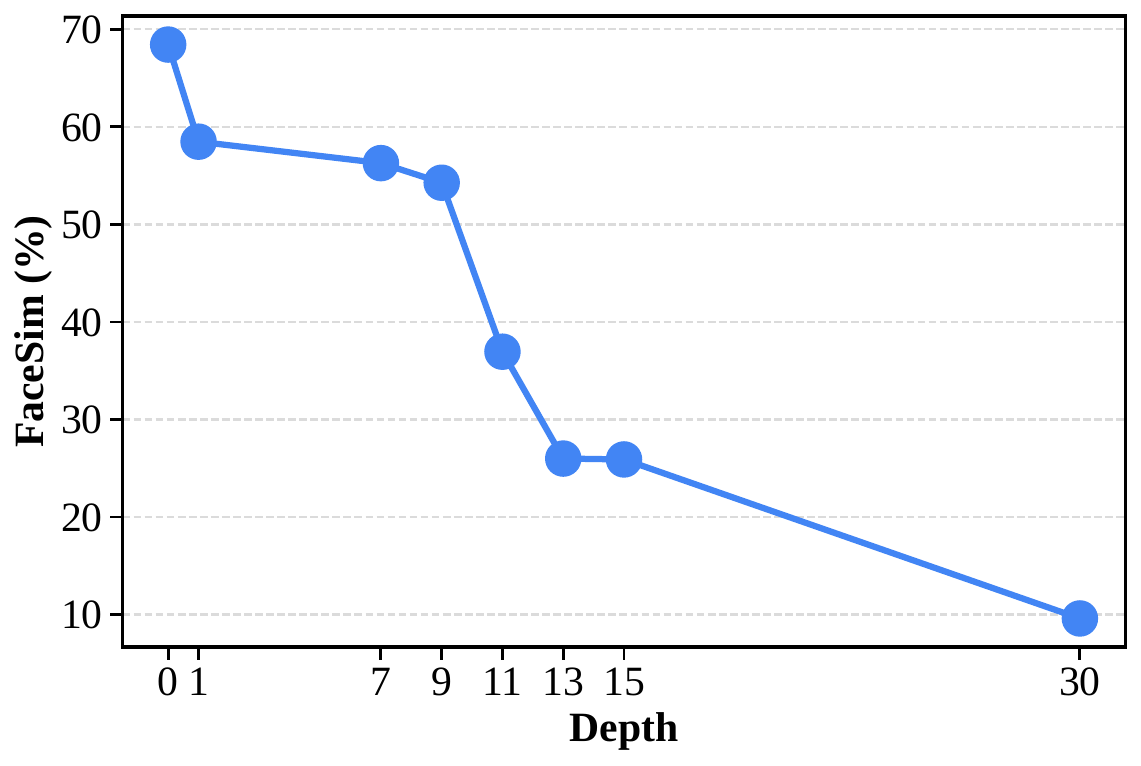}
        \vspace{-0.5cm}
        \caption{}
        \label{fig:depth_facesim}
    \end{subfigure}\hfill
    \begin{subfigure}[t]{0.32\columnwidth}
        \centering
        \includegraphics[width=\linewidth]{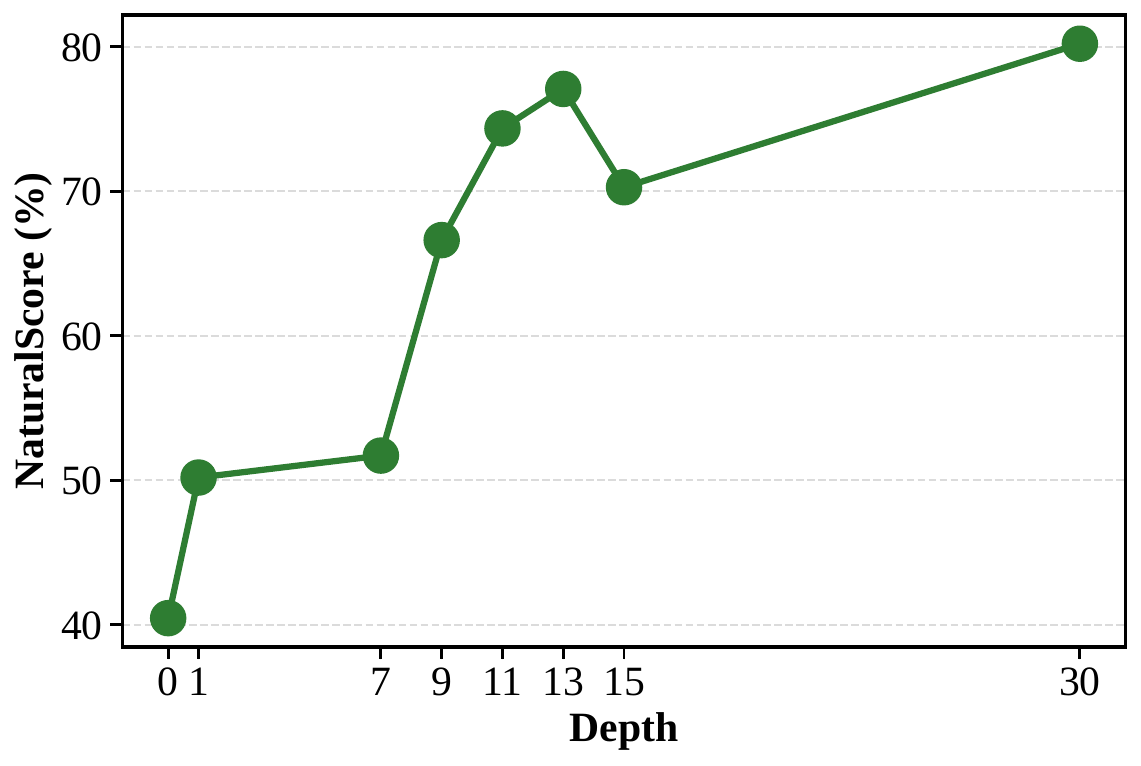}
        \vspace{-0.5cm}
        \caption{}
        \label{fig:depth_naturalscore}
    \end{subfigure}
    % \vspace{-0.3cm}
    \caption{Effect of alignment depth on three key metrics. Subfigures (a), (b), and (c) illustrate how alignment depth affects TotalScore, FaceSim, and NaturalScore, respectively. We treat w/o $\mathcal{L}_{RA}$ as depth 0.}
    \label{fig:depth_ablation_curves}
    % \vspace{-0.3cm}
\end{figure}

\noindent\textbf{Effectiveness of alignment depth.} To study the effect of alignment depth for the RA loss, we treat w/o $\mathcal{L}_{\mathrm{RA}}$ as depth $=0$ and compare different depths. As shown in Fig.~\ref{fig:depth_ablation_curves}~(\subref{fig:depth_totalscore}), TotalScore follows an ``increase-then-decrease'' trend and peaks at depth $=9$, suggesting that moderate depth yields a favorable trade-off. FaceSim decreases monotonically with depth (Fig.~\ref{fig:depth_ablation_curves}~(\subref{fig:depth_facesim})), indicating that overly deep alignment weakens subject identity fidelity, while NaturalScore generally increases with minor fluctuations (Fig.~\ref{fig:depth_ablation_curves}~(\subref{fig:depth_naturalscore})), indicating that deeper alignment improves naturalness. Based on this observation, we set depth $=9$ by default.

\noindent\textbf{Effectiveness of alignment encoder.} To evaluate the impact of the alignment encoder on the RA loss, we ablate encoder size and type. As shown in Tab.~\ref{tab:encoder_metrics}, using an alignment encoder consistently outperforms w/o $\mathcal{L}_{\mathrm{RA}}$ under both settings, suggesting the benefit of external representations as RA targets. \textbf{In the size ablation}, DINOv3-B/L/H+ all bring substantial gains, with only 0.33\% to 0.43\% variation in TotalScore, indicating that RA loss is robust to encoder size. Considering the trade-off between performance and computational cost, we use DINOv3-L as the default. \textbf{In the type ablation}, the improvement generalizes across different encoders (DINOv3-L, SigLIP2-So, and Qwen2.5-VL-7B), indicating it is not tied to a specific architecture. They also show distinct metric preferences: DINOv3-L performs best on consistency-related metrics (e.g., NexusScore and FaceSim), SigLIP2-So favors quality and naturalness metrics (e.g., Aesthetics, MotionSmoothness, and NaturalScore), while Qwen2.5-VL-7B performs better on MotionAmplitude. Overall, DINOv3-L achieves a favorable trade-off.

\begin{table}[t]
\centering
% \small
\caption{Ablation study on the impact of the alignment encoder in RefAlign on the OpenS2V-Eval.}
\label{tab:encoder_metrics}
\resizebox{0.999\linewidth}{!}{
\begin{tabular}{lcccccccc}
\midrule
Encoder & TotalScore$\uparrow$ & Aesthetics$\uparrow$ & MotionSmoothness$\uparrow$ & MotionAmplitude$\uparrow$ & FaceSim$\uparrow$ & GmeScore$\uparrow$ & NexusScore$\uparrow$ & NaturalScore$\uparrow$ \\
\midrule
w/o $\mathcal{L}_{\mathrm{RA}}$  & 49.93\% & 36.46\% & 88.97\% & 28.62\% & 68.45\% & 58.83\% & 38.63\% & 40.46\% \\
\midrule
\multicolumn{9}{c}{\textbf{\textit{Encoder size ablation}}} \\
\midrule
DINOv3-B  & 55.40\% & 38.83\% & 93.04\% & 22.49\% & 54.28\% & 66.07\% & 41.92\% & 66.62\% \\
DINOv3-L  & \textbf{55.73\%} & 41.53\% & 90.00\% & 29.53\% & 53.15\% & 67.54\% & 46.23\% & 62.96\% \\
DINOv3-H+ & 55.30\% & 43.72\% & 94.25\% & 20.09\% & 53.07\% & 66.90\% & 44.24\% & 61.48\% \\
\midrule
\multicolumn{9}{c}{\textbf{\textit{Encoder type ablation}}} \\
\midrule
DINOv3-L  & \textbf{55.73\%} & 41.53\% & 90.00\% & 29.53\% & 53.15\% & 67.54\% & 46.23\% & 62.96\% \\
SigLIP2-So  & 54.25\% & 43.95\% & 94.79\% & 17.90\% & 46.00\% & 68.43\% & 40.81\% & 65.00\% \\
Qwen2.5-VL-7B & 53.12\% & 37.50\% & 92.09\% & 32.13\% & 44.96\% & 64.40\% & 45.07\% & 63.43\% \\
\midrule
\end{tabular}
}
\end{table}

\subsection{User Study}

\begin{wrapfigure}{r}{0.50\textwidth}
  \centering
  \vspace{-1.5\baselineskip} 
  \includegraphics[width=0.50\textwidth]{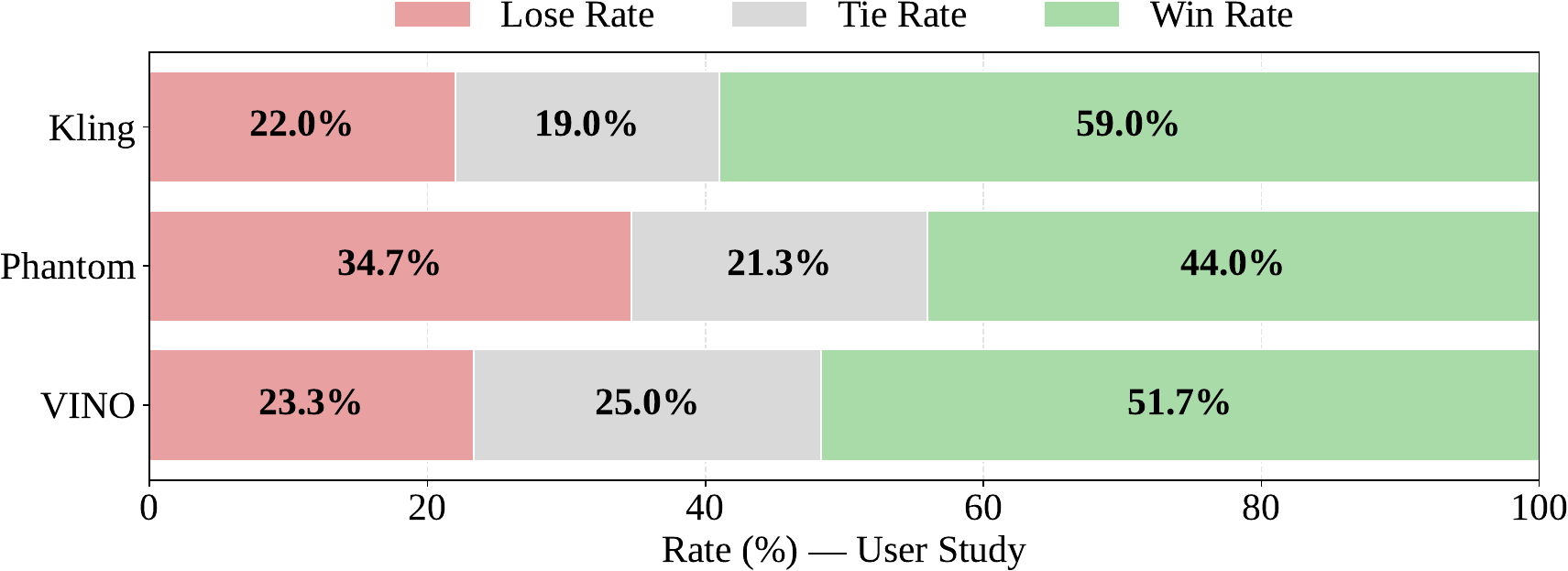}
  \caption{User study results comparing RefAlign with Kling, Phantom, and VINO.}
  \label{fig:human_pref}
  \vspace{-0.5\baselineskip} 
\end{wrapfigure}

We conduct a user study with 30 volunteers evaluating visual quality, reference fidelity, and video–text alignment. For each comparison pair, two videos from different methods are shown anonymously in random order, and participants choose the better one or a tie. As shown in Fig.~\ref{fig:human_pref}, RefAlign achieves the highest overall human preference.

\section{Conclusion}

This paper presents RefAlign, a representation alignment framework to enhance controllability in R2V generation. RefAlign aligns the reference-branch representation to an underlying vision foundation model (VFM) at the feature level via a reference alignment (RA) loss, helping alleviate copy--paste artifacts and multi-subject confusion from modality mismatch. The alignment is applied only during training and removed at inference, incurring no inference-time overhead. Extensive experiments show that RefAlign outperforms SOTA methods (e.g., Kling1.6, Phantom, and VINO) in reference subject fidelity and consistency. We hope this work provides new insights into reference-driven video generation and video editing, and further may promote a unified modeling paradigm for visual understanding and generation.

\noindent\textbf{Limitations and future work.} Limited training data diversity currently prevents an optimal balance between instruction following and reference fidelity. Moreover, the underlying foundation model limits RefAlign to 81-frame videos, restricting long-video generation. In addition, a single VFM guidance signal may be incomplete, as different VFM features emphasize different aspects of visual representations. Future work includes using more diverse data to improve the trade-off, extending R2V to longer videos, and combining multiple VFM features as alignment targets for more robust and comprehensive alignment.

\section*{Acknowledgement}
This work was supported by the National Natural Science Foundation of China under Grant Nos. 62361166670 and U24A20330, and by the Supercomputing Center of Nankai University (NKSC).

\appendix

\section{Additional Training Details}
During training, we randomly drop the prompt, the reference, or both, each with a probability of 10\%, for CFG. We train our model in two stages. In Stage 1, we use 200K regular-pair samples from OpenS2V~\cite{yuan2025opens2v} to better learn reference conditioning. In Stage 2, we use 160K cross-pair samples from Phantom-Data~\cite{chen2025phantom-data} to alleviate copy--paste artifacts.

\section{Additional Ablation Studies}
We present the detailed results of the alignment depth study in Tab.~\ref{tab:depth}.
\begin{table}[t]
\centering
% \small
\caption{Ablation study on the impact of alignment depth on OpenS2V-Eval~\cite{yuan2025opens2v}, with the best results in \textbf{bold}.}
\resizebox{0.999\linewidth}{!}{
\begin{tabular}{ccccccccc}
\toprule
\textbf{Depth} & \textbf{TotalScore}$\uparrow$ & \textbf{Aesthetics}$\uparrow$ & \textbf{MotionSmoothness}$\uparrow$ & \textbf{MotionAmplitude}$\uparrow$ & \textbf{FaceSim}$\uparrow$ & \textbf{GmeScore}$\uparrow$ & \textbf{NexusScore}$\uparrow$ & \textbf{NaturalScore}$\uparrow$ \\
\midrule
w/o $\mathcal{L}_{\mathrm{RA}}$  & 49.93\% & 36.46\% & 88.97\% & 28.62\% & 68.45\% & 58.83\% & 38.63\% & 40.46\% \\
1  & 52.32\% & 35.43\% & 92.96\% & 18.64\% & 58.49\% & 60.41\% & 48.53\% & 50.19\% \\
7  & 52.66\% & 41.39\% & 97.87\% & 7.78\% & 56.29\% & 62.15\% & 44.40\% & 51.71\% \\
9  & \textbf{55.40}\% & 38.83\% & 93.04\% & 22.49\% & 54.28\% & 66.07\% & 41.92\% & 66.62\% \\
11 & 53.87\% & 37.22\% & 90.93\% & 23.31\% & 36.95\% & 68.39\% & 42.78\% & 74.35\% \\
13 & 51.99\% & 35.45\% & 90.70\% & 23.15\% & 25.99\% & 67.98\% & 42.80\% & 77.08\% \\
15 & 51.35\% & 39.07\% & 90.12\% & 34.93\% & 25.90\% & 68.38\% & 43.71\% & 70.28\% \\
30 & 50.12\% & 32.57\% & 94.40\% & 17.98\% & 9.59\% & 68.86\% & 47.25\% & 80.21\% \\
\bottomrule
\end{tabular}
}
\label{tab:depth}
\end{table}

\section{Choice of the Vision Foundation Model}
We choose DINOv3~\cite{simeoni2025dinov3}, SigLIP2~\cite{tschannen2025siglip}, and Qwen2.5-VL-7B~\cite{bai2025qwen2} as the feature encoders for alignment targets due to their flexibility in handling non-square input resolutions. Our reference images have a resolution of 480×832, whereas many commonly used encoders (e.g., DINOv2~\cite{oquab2024dinov2}, CLIP~\cite{radford2021learning}, and MAE~\cite{he2022masked}) are trained with fixed square inputs such as 224×224, making them less suitable for non-square images. In contrast, DINOv3 employs RoPE~\cite{su2024roformer} and naturally supports arbitrary resolutions, while SigLIP2 and Qwen2.5-VL also accommodate flexible input resolutions and aspect ratios.

\section{User study details}
\begin{figure}[htbp]
    \centering
    \includegraphics[width=1.0\textwidth]{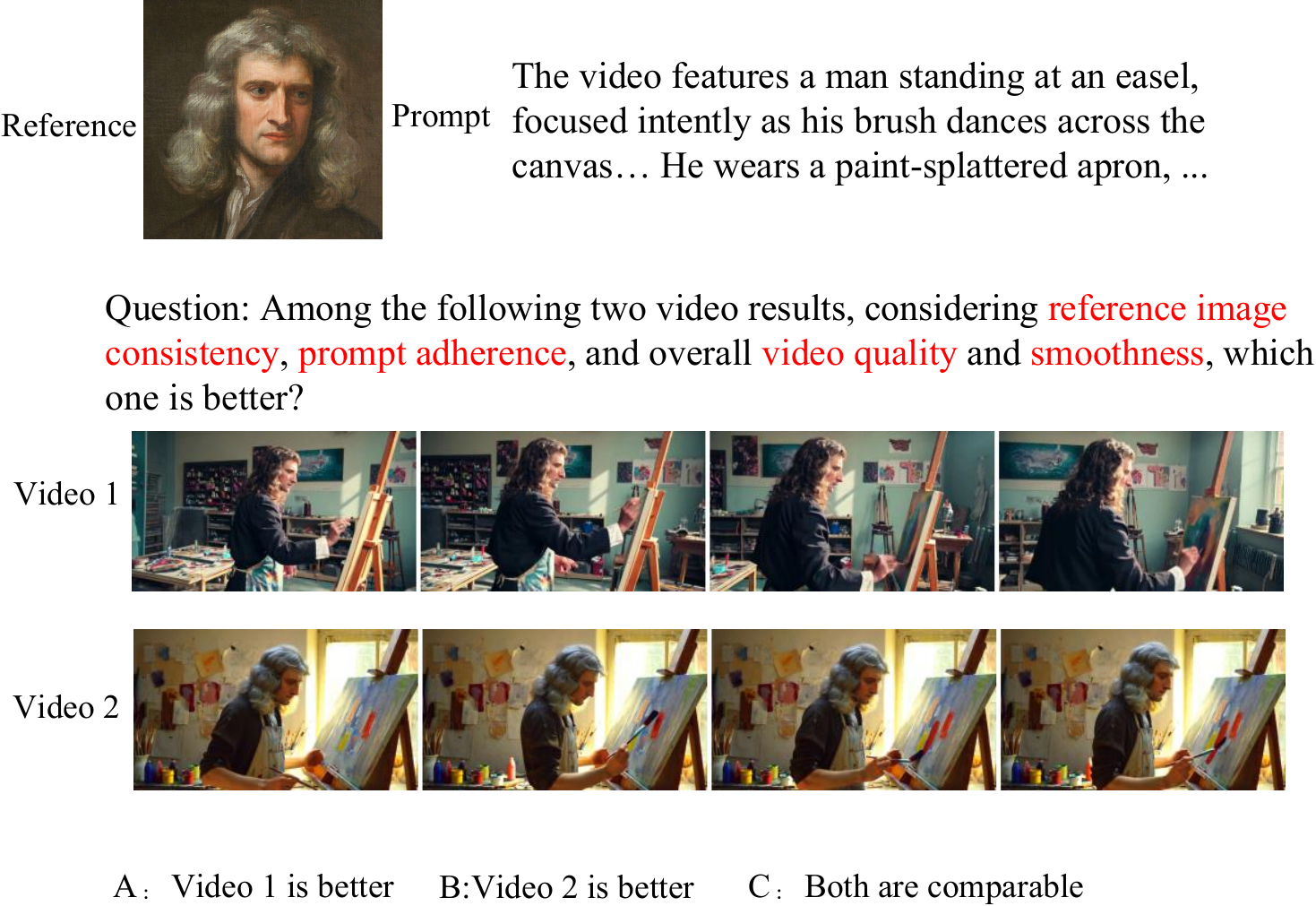}
    \caption{An example of the user study.}
    \label{fig:example}
\end{figure}
An example of the questionnaire is shown in the Fig~\ref{fig:example}.

\section{More Visualization Results}

\begin{figure}[htbp]
    \centering
    \includegraphics[width=0.68\textwidth]{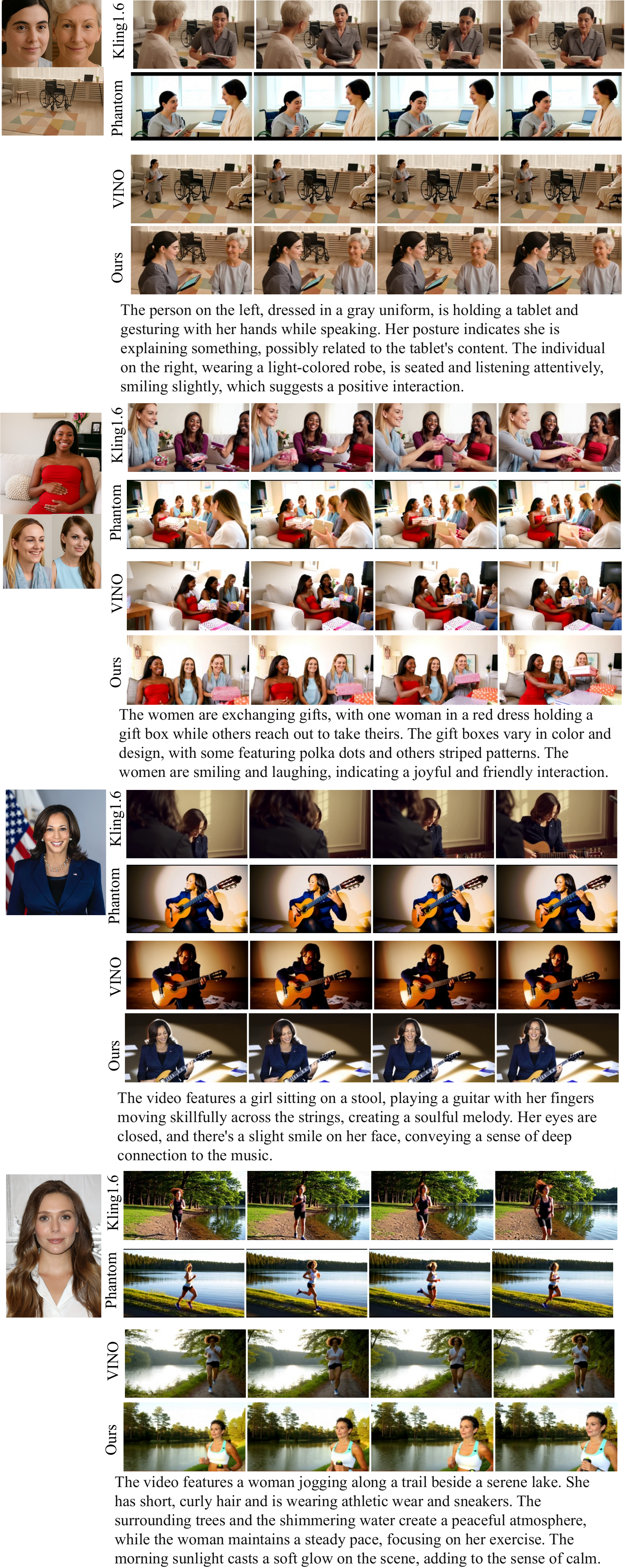}
    \caption{Qualitative results. We compare RefAlign with three representative methods,
namely Kling1.6~\cite{Kling}, Phantom~\cite{Liu_2025_ICCV}, and VINO~\cite{chen2026vino}.}
    \label{fig:maore_vis}
\end{figure}
\begin{figure}[htbp]
    \centering
    \includegraphics[width=\textwidth]{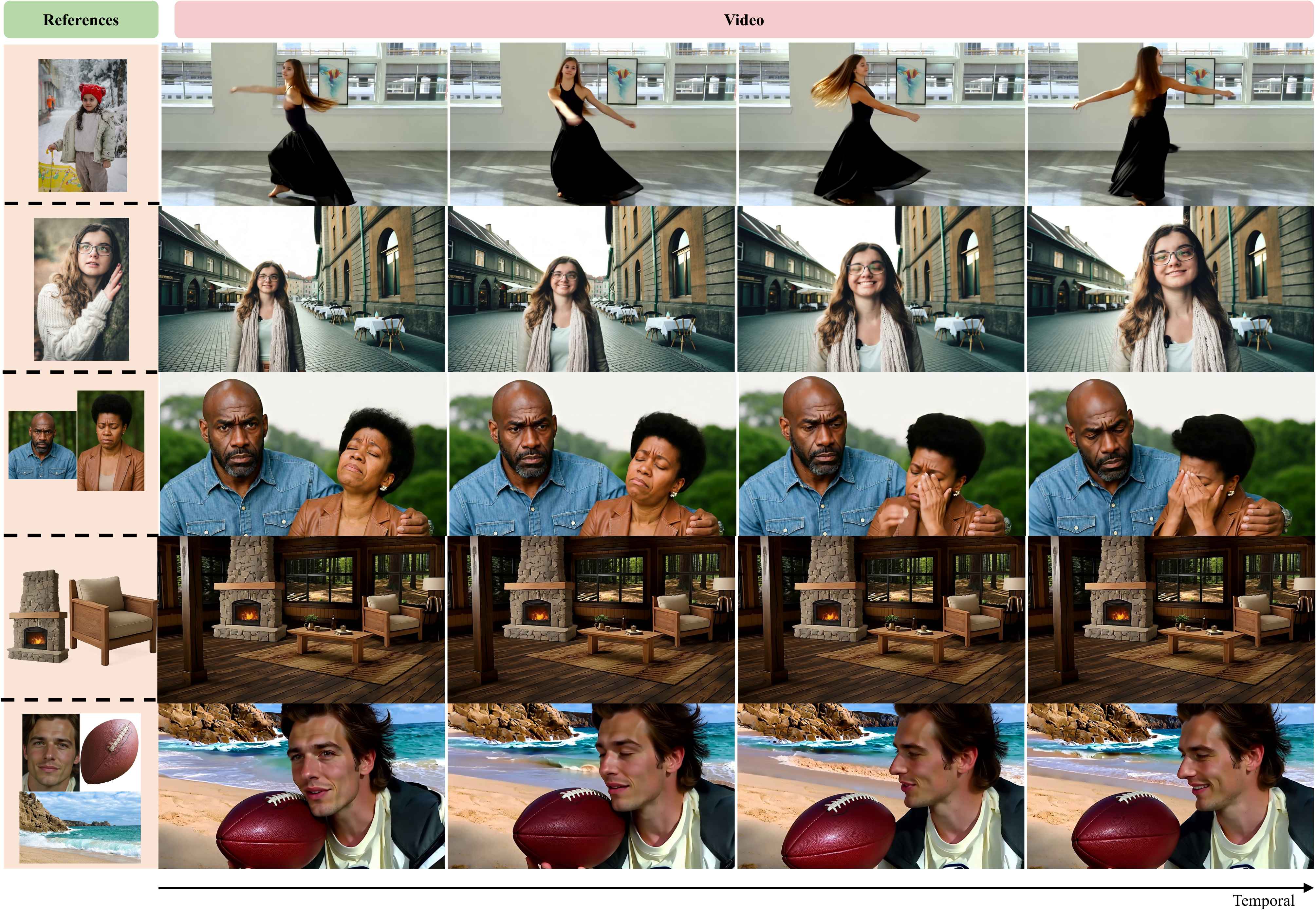}
    \caption{Reference-to-video generation using our proposed method, RefAlign.}
    \label{fig:best-vis}
\end{figure}

Additional visualization results of R2V generation are shown in the Fig.~\ref{fig:maore_vis} and Fig.~\ref{fig:best-vis}.

\bibliographystyle{splncs04}
\bibliography{main}

@String(CVPR  = {IEEE Conf. Comput. Vis. Pattern Recog.})

@String(ICCV  = {Int. Conf. Comput. Vis.})

@String(NeurIPS = {Adv. Neural Inform. Process. Syst.})

@String(ICML  = {Int. Conf. Mach. Learn.})

@String(ICLR  = {Int. Conf. Learn. Represent.})

@String(AAAI  = {AAAI})

@String(JMLR  = {J. Mach. Learn. Res.})

@String(TMLR  = {Trans. Mach. Learn Res.})

@String(CVPR  = {CVPR})

@String(ICCV  = {ICCV})

@String(NeurIPS = {NeurIPS})

@String(ICML  = {ICML})

@String(ICLR  = {ICLR})

@String(JMLR  = {JMLR})

@String(TMLR  = {TMLR})

@article{brooks2024video,
  title={Video generation models as world simulators},
  author={Brooks, Tim and Peebles, Bill and Holmes, Connor and DePue, Will and Guo, Yufei and Jing, Li and Schnurr, David and Taylor, Joe and Luhman, Troy and Luhman, Eric and others},
  journal={OpenAI Blog},
  volume={1},
  number={8},
  pages={1},
  year={2024}
}

@article{bao2024vidu,
  title={Vidu: a highly consistent, dynamic and skilled text-to-video generator with diffusion models},
  author={Bao, Fan and Xiang, Chendong and Yue, Gang and He, Guande and Zhu, Hongzhou and Zheng, Kaiwen and Zhao, Min and Liu, Shilong and Wang, Yaole and Zhu, Jun},
  journal={arXiv preprint arXiv:2405.04233},
  year={2024}
}

@article{team2025kling,
  title={Kling-Omni Technical Report},
  author={Team, Kling and Chen, Jialu and Ci, Yuanzheng and Du, Xiangyu and Feng, Zipeng and Gai, Kun and Guo, Sainan and Han, Feng and He, Jingbin and He, Kang and others},
  journal={arXiv preprint arXiv:2512.16776},
  year={2025}
}

@article{wan2025wan,
  title={Wan: Open and advanced large-scale video generative models},
  author={Wan, Team and Wang, Ang and Ai, Baole and Wen, Bin and Mao, Chaojie and Xie, Chen-Wei and Chen, Di and Yu, Feiwu and Zhao, Haiming and Yang, Jianxiao and others},
  journal={arXiv preprint arXiv:2503.20314},
  year={2025}
}

@inproceedings{yangcogvideox,
  title={CogVideoX: Text-to-Video Diffusion Models with An Expert Transformer},
  author={Yang, Zhuoyi and Teng, Jiayan and Zheng, Wendi and Ding, Ming and Huang, Shiyu and Xu, Jiazheng and Yang, Yuanming and Hong, Wenyi and Zhang, Xiaohan and Feng, Guanyu and others},
  booktitle=ICLR,
year={2025}
}

@article{wu2025hunyuanvideo,
  title={Hunyuanvideo 1.5 technical report},
  author={Wu, Bing and Zou, Chang and Li, Changlin and Huang, Duojun and Yang, Fang and Tan, Hao and Peng, Jack and Wu, Jianbing and Xiong, Jiangfeng and Jiang, Jie and others},
  journal={arXiv preprint arXiv:2511.18870},
  year={2025}
}

@inproceedings{chen2025multi,
  title={Multi-subject open-set personalization in video generation},
  author={Chen, Tsai-Shien and Siarohin, Aliaksandr and Menapace, Willi and Fang, Yuwei and Lee, Kwot Sin and Skorokhodov, Ivan and Aberman, Kfir and Zhu, Jun-Yan and Yang, Ming-Hsuan and Tulyakov, Sergey},
  booktitle=CVPR,
  pages={6099--6110},
  year={2025}
}

@article{huang2025conceptmaster,
  title={Conceptmaster: Multi-concept video customization on diffusion transformer models without test-time tuning},
  author={Huang, Yuzhou and Yuan, Ziyang and Liu, Quande and Wang, Qiulin and Wang, Xintao and Zhang, Ruimao and Wan, Pengfei and Zhang, Di and Gai, Kun},
  journal={arXiv preprint arXiv:2501.04698},
  year={2025}
}

@InProceedings{Liu_2025_ICCV,
    author    = {Liu, Lijie and Ma, Tianxiang and Li, Bingchuan and Chen, Zhuowei and Liu, Jiawei and Li, Gen and Zhou, Siyu and He, Qian and Wu, Xinglong},
    title     = {Phantom: Subject-Consistent Video Generation via Cross-Modal Alignment},
    booktitle = ICCV,
    month     = {October},
    year      = {2025},
    pages     = {14951-14961}
}

@inproceedings{nguyen2025swifttry,
  title={Swifttry: Fast and consistent video virtual try-on with diffusion models},
  author={Nguyen, Hung and Nguyen, Quang Qui-Vinh and Nguyen, Khoi and Nguyen, Rang},
  booktitle=AAAI,
  volume={39},
  pages={6200--6208},
  year={2025}
}

@inproceedings{li2025pursuing,
  title={Pursuing Temporal-Consistent Video Virtual Try-On via Dynamic Pose Interaction},
  author={Li, Dong and Zhong, Wenqi and Yu, Wei and Pan, Yingwei and Zhang, Dingwen and Yao, Ting and Han, Junwei and Mei, Tao},
  booktitle=CVPR,
  pages={22648--22657},
  year={2025}
}

@inproceedings{chen2025goku,
  title={Goku: Flow based video generative foundation models},
  author={Chen, Shoufa and Ge, Chongjian and Zhang, Yuqi and Zhang, Yida and Zhu, Fengda and Yang, Hao and Hao, Hongxiang and Wu, Hui and Lai, Zhichao and Hu, Yifei and others},
  booktitle=CVPR,
  pages={23516--23527},
  year={2025}
}

@inproceedings{liang2025movie,
  title={Movie Weaver: Tuning-Free Multi-Concept Video Personalization with Anchored Prompts},
  author={Liang, Feng and Ma, Haoyu and He, Zecheng and Hou, Tingbo and Hou, Ji and Li, Kunpeng and Dai, Xiaoliang and Juefei-Xu, Felix and Azadi, Samaneh and Sinha, Animesh and others},
  booktitle=CVPR,
  pages={13146--13156},
  year={2025}
}

@inproceedings{esser2024scaling,
  title={Scaling rectified flow transformers for high-resolution image synthesis},
  author={Esser, Patrick and Kulal, Sumith and Blattmann, Andreas and Entezari, Rahim and M{\"u}ller, Jonas and Saini, Harry and Levi, Yam and Lorenz, Dominik and Sauer, Axel and Boesel, Frederic and others},
  booktitle=ICML,
  year={2024}
}

@inproceedings{yurepresentation,
  title={Representation Alignment for Generation: Training Diffusion Transformers Is Easier Than You Think},
  author={Yu, Sihyun and Kwak, Sangkyung and Jang, Huiwon and Jeong, Jongheon and Huang, Jonathan and Shin, Jinwoo and Xie, Saining},
  booktitle=ICLR,
  year={2025}
}

@article{oquab2024dinov2,
  title={DINOv2: Learning Robust Visual Features without Supervision},
  author={Oquab, Maxime and Darcet, Timoth{\'e}e and Moutakanni, Th{\'e}o and Vo, Huy and Szafraniec, Marc and Khalidov, Vasil and Fernandez, Pierre and Haziza, Daniel and Massa, Francisco and El-Nouby, Alaaeldin and others},
  journal=TMLR,
  year={2024}
}

@article{raffel2020exploring,
  title={Exploring the limits of transfer learning with a unified text-to-text transformer},
  author={Raffel, Colin and Shazeer, Noam and Roberts, Adam and Lee, Katherine and Narang, Sharan and Matena, Michael and Zhou, Yanqi and Li, Wei and Liu, Peter J},
  journal=JMLR,
  volume={21},
  number={140},
  pages={1--67},
  year={2020}
}

@misc{Pika,
	author = "Pika",
	title = "Pikascenes",
	howpublished = "\url{https://pika.art/ingredients/}",
	year = "2024"
}

@misc{Kling,
	author = "kling",
	title = "Image to video elements feature",
	howpublished = "\url{https://klingai.com/image-to-video/multi-id/new/}",
	year = "2024"
}

@inproceedings{zhou2025scaling,
  title={Scaling zero-shot reference-to-video generation},
  author={Zhou, Zijian and Liu, Shikun and Liu, Haozhe and Qiu, Haonan and An, Zhaochong and Ren, Weiming and Liu, Zhiheng and Huang, Xiaoke and Ng, Kam-Woh and Xie, Tian and others},
  booktitle=CVPR,
  pages={9253--9262},
  year={2026}
}

@inproceedings{jiang2025vace,
  title={Vace: All-in-one video creation and editing},
  author={Jiang, Zeyinzi and Han, Zhen and Mao, Chaojie and Zhang, Jingfeng and Pan, Yulin and Liu, Yu},
  booktitle=ICCV,
  pages={17191--17202},
  year={2025}
}

@article{fei2025skyreels,
  title={Skyreels-a2: Compose anything in video diffusion transformers},
  author={Fei, Zhengcong and Li, Debang and Qiu, Di and Wang, Jiahua and Dou, Yikun and Wang, Rui and Xu, Jingtao and Fan, Mingyuan and Chen, Guibin and Li, Yang and others},
  journal={arXiv preprint arXiv:2504.02436},
  year={2025}
}

@article{deng2025magref,
  title={Magref: Masked guidance for any-reference video generation},
  author={Deng, Yufan and Guo, Xun and Yin, Yuanyang and Zhiyuan Fang, Jacob and Yang, Yiding and Wang, Yizhi and Yuan, Shenghai and Wang, Angtian and Liu, Bo and Huang, Haibin and others},
  journal=ICLR,
  year={2026}
}

@article{chen2026vino,
  title={VINO: A Unified Visual Generator with Interleaved OmniModal Context},
  author={Chen, Junyi and He, Tong and Fu, Zhoujie and Wan, Pengfei and Gai, Kun and Ye, Weicai},
  journal={arXiv preprint arXiv:2601.02358},
  year={2026}
}

@article{li2025bindweave,
  title={Bindweave: Subject-consistent video generation via cross-modal integration},
  author={Li, Zhaoyang and Qian, Dongjun and Su, Kai and Diao, Qishuai and Xia, Xiangyang and Liu, Chang and Yang, Wenfei and Zhang, Tianzhu and Yuan, Zehuan},
  journal=ICLR,
  year={2026}
}

@article{zhang2025kaleido,
  title={Kaleido: Open-Sourced Multi-Subject Reference Video Generation Model},
  author={Zhang, Zhenxing and Teng, Jiayan and Yang, Zhuoyi and Cao, Tiankun and Wang, Cheng and Gu, Xiaotao and Tang, Jie and Guo, Dan and Wang, Meng},
  journal={arXiv preprint arXiv:2510.18573},
  year={2025}
}

@article{xue2025stand,
  title={Stand-in: A lightweight and plug-and-play identity control for video generation},
  author={Xue, Bowen and Duan, Zheng-Peng and Yan, Qixin and Wang, Wenjing and Liu, Hao and Guo, Chun-Le and Li, Chongyi and Li, Chen and Lyu, Jing},
  journal=CVPR,
  year={2026}
}

@inproceedings{zhong2025concat,
  title={Concat-ID: Towards Universal Identity-Preserving Video Synthesis},
  author={Zhong, Yong and Yang, Zhuoyi and Teng, Jiayan and Gu, Xiaotao and Li, Chongxuan},
  booktitle={ICCVW},
  pages={1906--1915},
  year={2025}
}

@inproceedings{yuan2025identity,
  title={Identity-preserving text-to-video generation by frequency decomposition},
  author={Yuan, Shenghai and Huang, Jinfa and He, Xianyi and Ge, Yunyang and Shi, Yujun and Chen, Liuhan and Luo, Jiebo and Yuan, Li},
  booktitle=CVPR,
  pages={12978--12988},
  year={2025}
}

@inproceedings{sang2025lynx,
  title    = {Lynx: Towards High-Fidelity Personalized Video Generation},
  author   = {Sang, Shen and Zhi, Tiancheng and Gu, Tianpei and Liu, Jing and Luo, Linjie},
  booktitle=CVPR,
  year={2026}
}

@article{liu2023visual,
  title={Visual instruction tuning},
  author={Liu, Haotian and Li, Chunyuan and Wu, Qingyang and Lee, Yong Jae},
  journal=NeurIPS,
  volume={36},
  pages={34892--34916},
  year={2023}
}

@article{bai2025qwen2,
  title={Qwen2. 5-vl technical report},
  author={Bai, Shuai and Chen, Keqin and Liu, Xuejing and Wang, Jialin and Ge, Wenbin and Song, Sibo and Dang, Kai and Wang, Peng and Wang, Shijie and Tang, Jun and others},
  journal={arXiv preprint arXiv:2502.13923},
  year={2025}
}

@article{hu2025polyvivid,
  title={PolyVivid: Vivid Multi-Subject Video Generation with Cross-Modal Interaction and Enhancement},
  author={Hu, Teng and Yu, Zhentao and Zhou, Zhengguang and Zhang, Jiangning and Zhou, Yuan and Lu, Qinglin and Yi, Ran},
  journal=NeurIPS,
  year={2025}
}

@article{deng2025cinema,
  title={Cinema: Coherent multi-subject video generation via mllm-based guidance},
  author={Deng, Yufan and Guo, Xun and Wang, Yizhi and Fang, Jacob Zhiyuan and Wang, Angtian and Yuan, Shenghai and Yang, Yiding and Liu, Bo and Huang, Haibin and Ma, Chongyang},
  journal={arXiv preprint arXiv:2503.10391},
  year={2025}
}

@article{hu2025hunyuancustom,
  title={Hunyuancustom: A multimodal-driven architecture for customized video generation},
  author={Hu, Teng and Yu, Zhentao and Zhou, Zhengguang and Liang, Sen and Zhou, Yuan and Lin, Qin and Lu, Qinglin},
  journal={arXiv preprint arXiv:2505.04512},
  year={2025}
}

@article{pan2025id,
  title={ID-Crafter: VLM-Grounded Online RL for Compositional Multi-Subject Video Generation},
  author={Pan, Panwang and Zhao, Jingjing and Lin, Yuchen and Lin, Chenguo and Li, Chenxin and Liu, Hengyu and Shen, Tingting and Mu, Yadong},
  journal=CVPR,
  year={2026}
}

@inproceedings{leng2025repa,
  title={Repa-e: Unlocking vae for end-to-end tuning of latent diffusion transformers},
  author={Leng, Xingjian and Singh, Jaskirat and Hou, Yunzhong and Xing, Zhenchang and Xie, Saining and Zheng, Liang},
  booktitle=ICCV,
  pages={18262--18272},
  year={2025}
}

@article{wang2025ddt,
  title={Ddt: Decoupled diffusion transformer},
  author={Wang, Shuai and Tian, Zhi and Huang, Weilin and Wang, Limin},
  journal=CVPR,
  year={2026}
}

@article{wu2025representation,
  title={Representation Entanglement for Generation: Training Diffusion Transformers Is Much Easier Than You Think},
  author={Wu, Ge and Zhang, Shen and Shi, Ruijing and Gao, Shanghua and Chen, Zhenyuan and Wang, Lei and Chen, Zhaowei and Gao, Hongcheng and Tang, Yao and Yang, Jian and others},
  journal=NeurIPS,
  year={2025}
}

@article{kouzelis2025boosting,
  title={Boosting Generative Image Modeling via Joint Image-Feature Synthesis},
  author={Kouzelis, Theodoros and Karypidis, Efstathios and Kakogeorgiou, Ioannis and Gidaris, Spyros and Komodakis, Nikos},
  journal=NeurIPS,
  year={2025}
}

@inproceedings{yao2025reconstruction,
  title={Reconstruction vs. generation: Taming optimization dilemma in latent diffusion models},
  author={Yao, Jingfeng and Yang, Bin and Wang, Xinggang},
  booktitle=CVPR,
  pages={15703--15712},
  year={2025}
}

@article{xie2025unleashing,
  title={Unleashing the Potential of Large Language Models for Text-to-Image Generation through Autoregressive Representation Alignment},
  author={Xie, Xing and Liu, Jiawei and Lin, Ziyue and Fan, Huijie and Han, Zhi and Tang, Yandong and Qu, Liangqiong},
  journal=AAAI,
  year={2026}
}

@article{zhang2025videorepa,
  title={VideoREPA: Learning Physics for Video Generation through Relational Alignment with Foundation Models},
  author={Zhang, Xiangdong and Liao, Jiaqi and Zhang, Shaofeng and Meng, Fanqing and Wan, Xiangpeng and Yan, Junchi and Cheng, Yu},
  journal=NeurIPS,
  year={2025}
}

@inproceedings{ho2022classifier,
  title={Classifier-Free Diffusion Guidance},
  author={Ho, Jonathan and Salimans, Tim},
  booktitle={NeurIPS Workshop on Deep Generative Models and Downstream Applications},
  year={2021}
}

@article{loshchilov2017decoupled,
  title={Decoupled weight decay regularization},
  author={Loshchilov, Ilya and Hutter, Frank},
  journal=ICLR,
  year={2019}
}

@article{yuan2025opens2v,
  title={OpenS2V-Nexus: A Detailed Benchmark and Million-Scale Dataset for Subject-to-Video Generation},
  author={Yuan, Shenghai and He, Xianyi and Deng, Yufan and Ye, Yang and Huang, Jinfa and Lin, Bin and Luo, Jiebo and Yuan, Li},
  journal=NeurIPS,
  year={2025}
}

@article{chen2025phantom-data,
      title={Phantom-Data: Towards a General Subject-Consistent Video Generation Dataset},
      author={Chen, Zhuowei and Li, Bingchuan and Ma, Tianxiang and Liu, Lijie and Liu, Mingcong and Zhang, Yi and Li, Gen and Li, Xinghui and Zhou, Siyu and He, Qian and Wu, Xinglong},
      journal=ICLR,
      year={2026}
    }

@article{wang2024qwen2,
  title={Qwen2-vl: Enhancing vision-language model's perception of the world at any resolution},
  author={Wang, Peng and Bai, Shuai and Tan, Sinan and Wang, Shijie and Fan, Zhihao and Bai, Jinze and Chen, Keqin and Liu, Xuejing and Wang, Jialin and Ge, Wenbin and others},
  journal={arXiv preprint arXiv:2409.12191},
  year={2024}
}

@article{simeoni2025dinov3,
  title={Dinov3},
  author={Sim{\'e}oni, Oriane and Vo, Huy V and Seitzer, Maximilian and Baldassarre, Federico and Oquab, Maxime and Jose, Cijo and Khalidov, Vasil and Szafraniec, Marc and Yi, Seungeun and Ramamonjisoa, Micha{\"e}l and others},
  journal={arXiv preprint arXiv:2508.10104},
  year={2025}
}

@article{tschannen2025siglip,
  title={Siglip 2: Multilingual vision-language encoders with improved semantic understanding, localization, and dense features},
  author={Tschannen, Michael and Gritsenko, Alexey and Wang, Xiao and Naeem, Muhammad Ferjad and Alabdulmohsin, Ibrahim and Parthasarathy, Nikhil and Evans, Talfan and Beyer, Lucas and Xia, Ye and Mustafa, Basil and others},
  journal={arXiv preprint arXiv:2502.14786},
  year={2025}
}

@article{JMLR:v9:vandermaaten08a,
  author  = {Laurens van der Maaten and Geoffrey Hinton},
  title   = {Visualizing Data using t-SNE},
  journal = JMLR,
  year    = {2008},
  volume  = {9},
  number  = {86},
  pages   = {2579--2605},
  url     = {http://jmlr.org/papers/v9/vandermaaten08a.html}
}

@inproceedings{radford2021learning,
  title={Learning transferable visual models from natural language supervision},
  author={Radford, Alec and Kim, Jong Wook and Hallacy, Chris and Ramesh, Aditya and Goh, Gabriel and Agarwal, Sandhini and Sastry, Girish and Askell, Amanda and Mishkin, Pamela and Clark, Jack and others},
  booktitle=ICML,
  pages={8748--8763},
  year={2021},
  organization={PmLR}
}

@inproceedings{he2022masked,
  title={Masked autoencoders are scalable vision learners},
  author={He, Kaiming and Chen, Xinlei and Xie, Saining and Li, Yanghao and Doll{\'a}r, Piotr and Girshick, Ross},
  booktitle=CVPR,
  pages={16000--16009},
  year={2022}
}

@article{su2024roformer,
  title={Roformer: Enhanced transformer with rotary position embedding},
  author={Su, Jianlin and Ahmed, Murtadha and Lu, Yu and Pan, Shengfeng and Bo, Wen and Liu, Yunfeng},
  journal={Neurocomputing},
  volume={568},
  pages={127063},
  year={2024},
  publisher={Elsevier}
}
\end{document}